\documentclass{sig-alternate}[10pt]
\usepackage{mathptmx}
\usepackage{fancyhdr}
\usepackage{hyperref}
\usepackage[sort,nocompress]{cite}
\usepackage[final]{microtype}
\usepackage{flushend}
\usepackage[]{subfig}
\usepackage[table]{xcolor}
\usepackage{xspace}

\newcommand{\BASE}{\textit{DCNN}\xspace}
\newcommand{\STR}{\textit{STR}\xspace}
\newcommand{\PRA}{\textit{Pragmatic}\xspace}
\newcommand{\PRAS}{\textit{PRA}\xspace}
\newcommand{\OURcore}{Bit-Tactical}
\newcommand{\OURScore}{TCL}
\newcommand{\OURSTRcore}{TCLp}
\newcommand{\OURPRAcore}{TCLe}

\newcommand{\OUR}{\textit{\OURcore}\xspace}
\newcommand{\OURS}{\textit{\OURScore}\xspace}
\newcommand{\OURSTR}{\textit{\OURSTRcore}\xspace}
\newcommand{\OURPRA}{\textit{\OURPRAcore}\xspace}

\fancypagestyle{firstpage}{
  \fancyhf{}
  
  \pagenumbering{arabic}
}

\title{\OURcore: Exploiting Ineffectual Computations in Convolutional Neural Networks: Which, Why, and How}
\author{Alberto Delmas Lascorz, Patrick Judd, Dylan Malone Stuart, Zissis Poulos,\\ Mostafa Mahmoud, Sayeh Sharify, Milos Nikolic, Andreas Moshovos\\
Electrical and Computer Engineering, University of Toronto\\\{delmasl1, juddpatr, malones2, zpoulos, sayeh, moshovos\}@ece.utoronto.ca,\\ \{mostafa.mahmoud, milos.nikolic\}@mail.utoronto.ca}

\newcommand{\OURSTRperf}{$5.05\times$\xspace}
\newcommand{\OURSTReff}{$2.98\times$\xspace}
\newcommand{\OURSTRarea}{22\%\xspace}
\newcommand{\OURSTRareatimes}{$1.22\times$\xspace}

\newcommand{\OURPRAperf}{$8.67\times$\xspace}
\newcommand{\OURPRAeff}{$2.97\times$\xspace}
\newcommand{\OURPRAarea}{$1.47\times$\xspace}

\newcommand{\OURSTRlowperf}{$4.6\times$\xspace}

\newcommand{\SCNNvsOURSdense}{46\% slower\xspace}
\newcommand{\SCNNvsOURSresnet}{48\% slower\xspace}
\newcommand{\SCNNvsOURSTR}{$2.78\times$\xspace}
\newcommand{\SCNNvsOURPRA}{$4.77\times$\xspace}

\begin{document}
\maketitle
\thispagestyle{firstpage}

\begin{abstract}

We show that, during inference with Convolutional Neural Networks (CNNs), more than $2\times$ to $8\times$ ineffectual work can be exposed if instead of targeting those weights and activations that are zero, we target different combinations of value stream properties. We demonstrate a practical application with \OUR (\OURS), a hardware accelerator  which exploits weight sparsity, per layer precision variability and dynamic fine-grain precision reduction for activations, and optionally the naturally occurring sparse effectual bit content of activations to improve performance and energy efficiency. \OURS benefits both \textit{sparse} and \textit{dense} CNNs, natively supports both convolutional and fully-connected layers, and exploits properties of \textit{all} activations to reduce storage, communication, and computation demands. While \OURS does not require changes to the CNN to deliver benefits, it does reward any technique that would amplify any of the aforementioned weight and activation value properties. Compared to an equivalent data-parallel accelerator for dense CNNs, \OURSTR, a variant of \OURS improves performance by \OURSTRperf and is \OURSTReff more energy efficient while requiring \OURSTRarea more area.

\end{abstract}

\section{Introduction}
\label{sec:intro}
Deep Neural Networks (DNNs) have been finding a growing number of applications while they are being deployed over a variety of computing platforms from high-end servers to mobile and embedded systems. Given the growing importance of DNN workloads and their high computation and memory demands, specialized hardware accelerators have emerged. While a few different types of DNNs exist, Convolutional Neural Networks (CNNs) are dominating image-based applications. Of particular interest are applications where CNNs are used for inference of input images or video frames.
Accordingly, this work targets the acceleration of inference with CNNs and specifically, the acceleration of convolutional layers which tend to dominate execution time in CNNs~\cite{DaDiannao}.

Early successes in hardware acceleration for CNNs exploited primarily their computation structure and the significant reuse in their memory access stream, e.g.,~\cite{diannao,DaDiannao,isscc_2016_chen_eyeriss}. Many recent CNN hardware accelerators exploit the various forms of  \textit{informational inefficiency} that CNNs exhibit. Informational inefficiency manifests in CNNs as ineffectual weights~\cite{han_eie:isca_2016,CambriconXMICRO16,SCNN}, activations~\cite{han_eie:isca_2016,albericio:cnvlutin,SCNN}, as an excess of precision, e.g.,~\cite{kim_x1000_2014,judd:reduced,binaryconnect,quantizedBlog,DBLP:journals/corr/VenkateshNM16,Stripes-MICRO},  as ineffectual activation bits~\cite{Pragmatic}, or in general as over-provisioning, e.g.,~\cite{squeezenet}. A shared goal of these designs is to reduce the number of operations needed to execute a CNN.

Accelerators that target zero-valued weights (the corresponding computations can be safely skipped) have received significant attention due to advances in weight pruning techniques. For example, for the set of state-of-the-art image classification CNNs considered here (see Section~\ref{sec:e}), the fraction of remaining non-zero weights after pruning is as little as 10\% for AlexNet and up to 50\% for the more recent ResNet-50~\cite{han_deep_2015-1,cvpr_2017_yang_energy,SkimCaffePaper}. Respectively, the performance improvement potential is $10\times$ and $2\times$ by weight skipping alone. Among those accelerators, Cambricon-X targeted ineffectual weights~\cite{CambriconXMICRO16} whereas SCNN targeted also zero-valued activations~\cite{SCNN}. The performance potential from skipping also zero-valued activations nearly doubles as 40\% to 50\% of them tend to be zero~\cite{albericio:cnvlutin,EyerissISCA2016,Reagen2016,han_eie:isca_2016,SCNN}. While both designs aim to remove \textit{all} relevant ineffectual input values, other inefficiencies limit the final performance and energy-efficiency benefits. We corroborate these findings in Section~\ref{sec:e:scnn} for SCNN, the most aggressive and higher performing design.  Work imbalance across processing elements (PEs), per PE working set sizes, the use of full crossbars to steer individual products to the respective accumulator arrays, and inter-PE communication for partial output values are the primary culprits~\cite{SCNN}.

Given that other sources of hardware- and application-level inefficiency ultimately limit the performance benefits possible when skipping all zero values and given that other sources of ineffectual computations have been identified, this work revisits the question on \textit{which} ineffectual computations to exploit and \textit{how} to do so.
We show that the performance potential is higher (e.g., $2\times$ to $8\times$ for a pruned AlexNet) if instead of targeting the zero-valued activations, we target the dynamic precision requirements or the ineffectual bit content of \textit{all} activations. This suggests that a design that exploits zero weights and the precision requirements or the ineffectual bit content of activations has a lot more flexibility to sacrifice some of this performance potential to better balance complexity, area, performance and energy while still outperforming designs based on zero weight and activation skipping.

Motivated by these observations and to show that they can have a practical application we present \OUR, or \OURS for short, a hardware accelerator for CNN inference.  At the weight front, similarly to Cambricon-X and SCNN, \OURS takes advantage of weight sparsity~\cite{DynamiCNetworkSurgery,han_deep_2015-1}, if present, to avoid computations with ineffectual weights. Specifically, in \OURS ineffectual weight elimination is orchestrated statically by rearranging the weights into an appropriate schedule in advance of the CNN execution and via a software scheduler. \OURS supports a limited set of weight promotions judiciously sacrificing some of the scheduling flexibility and thus some of the performance potential \textit{upfront}. However, this enables \OURS to use small multiplexers (4- to 8- input multiplexers prove sufficient for our purposes).

At the activation front, \OURS does not target zero-valued activations explicitly but does so implicitly so that it can exploit ineffectual computations found in \textit{all} activations. Section~\ref{sec:motivate} demonstrates the potential of this approach quantitatively while Section~\ref{sec:ia:whynot} explains why this is so qualitatively.
We present two variants of \OURS that deliver different area, performance and energy efficiency trade offs. The first, \OURPRA,
exploits the lopsided distribution of effectual activation bits: On average in CNNs less than 10\% of the activation bits are one which means that on average at least 90\% of the work performed when multiplying activations is ineffectual~\cite{Pragmatic}.
The second variant, \OURSTR, exploits the variable precision requirements of activations to scale execution time proportionally with the number of precision bits the activations need. This variant benefits from profile-derived per layer precisions if available. Regardless, at runtime it further trims any pre-specified activation precisions at a much finer granularity and to the minimum necessary to represent the specific activation values at hand. \OURPRA offers higher performance compared to \OURSTR, however, \OURSTR requires less area and is more energy efficient.

\OURS \textit{simultaneously} exploits: 1)~weight sparsity,
2)~the sparse effectual bit content of activations, and 3)~the dynamic precision variability of precisions at the granularity of a compute block. We find that exploiting these properties results in nearly \textit{multiplicative} benefits.
Further, \OURS benefits the convolutional layers of both \textit{dense} and \textit{sparse} CNNs, while taking advantage of \textit{all} activations, \textit{ineffectual} and \textit{effectual}. Moreover, it naturally supports fully-connected layers matching the performance of a conventional accelerator for those layers.

We highlight the following contributions:
\begin{itemize}
\item We demonstrate that more ineffectual computations can be found if, instead of targeting ineffectual weights and activations, we choose to target ineffectual weights in conjunction with either ineffectual activation bits or the runtime precision requirements of activations.
\item We present a front-end weight skipping design that combines both software and hardware to avoid the area- and energy-expensive full-crossbars of past designs. The front-end uses a small multiplexer per weight (4- to 8-input) and per activation (2- to 4-input) resulting into an energy- and area-efficient interconnect. It relies on pre-scheduling of the statically known weights to make the most of the sparse-hardware connectivity to eliminate \textit{enough} ineffectual weights.
\item We present our design as an extension over a vector-like engine which uses adder trees to reduce several products into a single sum prior to storing into an accumulator. Zhang \textit{et al.,} note that it was not possible to extend a similar design to support weight skipping in order to motivate the Cambricon-X design~\cite{CambriconX}. We believe that this work serves as motivation to consider how our approach can be incorporated in grid-like designs such as the Eyeriss~\cite{EyerissISCA2016} or the TPU~\cite{TPUISCA17}.
\end{itemize}

Experimental results with sparse versions of popular image classification CNNs show that a modest configuration of \OURPRA  improves performance by \OURPRAperf and is \OURPRAeff more energy efficient compared to an equivalent dense CNN accelerator. In this configuration weights can advance only by up to 2 positions ahead or up to 5 positions to the side. It requires only \OURPRAarea more area than the equivalent dense CNN accelerator, assuming that it does not take advantage of reduced precisions to reduce on-chip storage.
An equivalent \OURSTR configuration is \OURSTRperf faster, \OURSTReff more energy efficient while requiring \OURSTRareatimes more area than the dense accelerator.
Finally, we compare with a design that targets ineffectual weights and activations.

The rest of this document is organized as follows: Section~\ref{sec:motivate} motivates this work by comparing the performance potential of targeting various ineffectual computations present in CNNs. Section~\ref{sec:bg} reviews the operation of convolutional layers in CNNs and presents the baseline accelerator that targets dense CNNs. Section~\ref{sec:ews} presents how \OUR eliminates ineffectual weights. Section~\ref{sec:ea} discusses \OURPRA and \OURSTR. Section~\ref{sec:e} evaluates the designs. Finally, Section~\ref{sec:related} discusses related work, Section~\ref{sec:limitations} discusses limitations, while Section~\ref{sec:theend} concludes.

\section{Motivation}
\label{sec:motivate}
\begin{figure}
\includegraphics[width=0.48\textwidth]{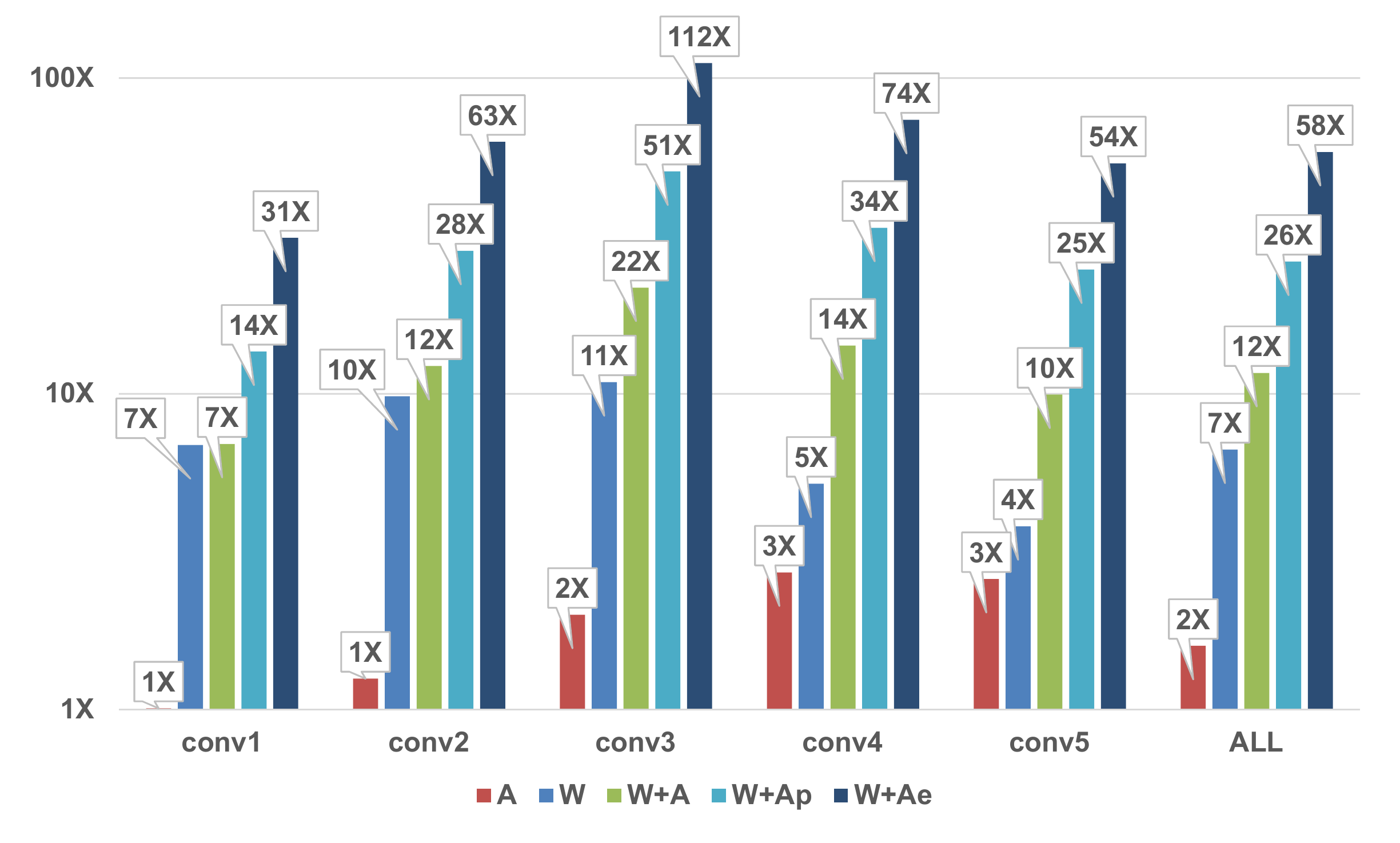}
\caption{AlexNet-ES: Performance improvement potential when removing ineffectual computations (logarithmic scale).}
\label{fig:motivate}
\end{figure}

Figure~\ref{fig:motivate} shows the fraction of total work that can ideally be removed, expressed as a relative speedup potential over a machine that performs all computations. It considers the following ineffectual computations: \textbf{A:}~zero activations, \textbf{W:}~zero weights, \textbf{W+A:}~zero weights and activations, \textbf{W+Ap:}~zero weights and the dynamic precision requirements of the activations, and \textbf{W+Ae:}~zero weights and the ineffectual bit content of the activations. For clarity the figure reports results only for AlexNet-ES~\cite{cvpr_2017_yang_energy}, a pruned version of AlexNet. The figure reports per layer results for the convolutional layers and the average across all those layers.

Whereas skipping weights or activations alone could improve performance on average by $6.7\times$ or $1.59\times$ respectively, skipping both boosts the potential to $11.6\times$. These results corroborate the findings of Parashar \textit{et al.}, that skipping zero-valued weights and activations has considerable performance potential~\cite{SCNN}. Furthermore, they show that for all layers most of the benefits are due to weight skipping.

However, compared to W+A,  W+Ap or W+Ae increase the average potential benefits by more than $2\times$ and $8\times$ respectively to $26\times$ and $58\times$. Moreover, this increase in potential is consistent across all layers. This observation suggests that
a design that targets W+Ap or W+Ae has a lot more flexibility to sacrifice some of the potential from either the W or the A side in order to better balance complexity, performance, area, and energy. It is this observation that underpins the \OURS design.

\section{Background}
\label{sec:bg}
\subsection{Convolutional Layers}
\label{sec:cvl}
\begin{figure}
\includegraphics[width=0.48\textwidth]{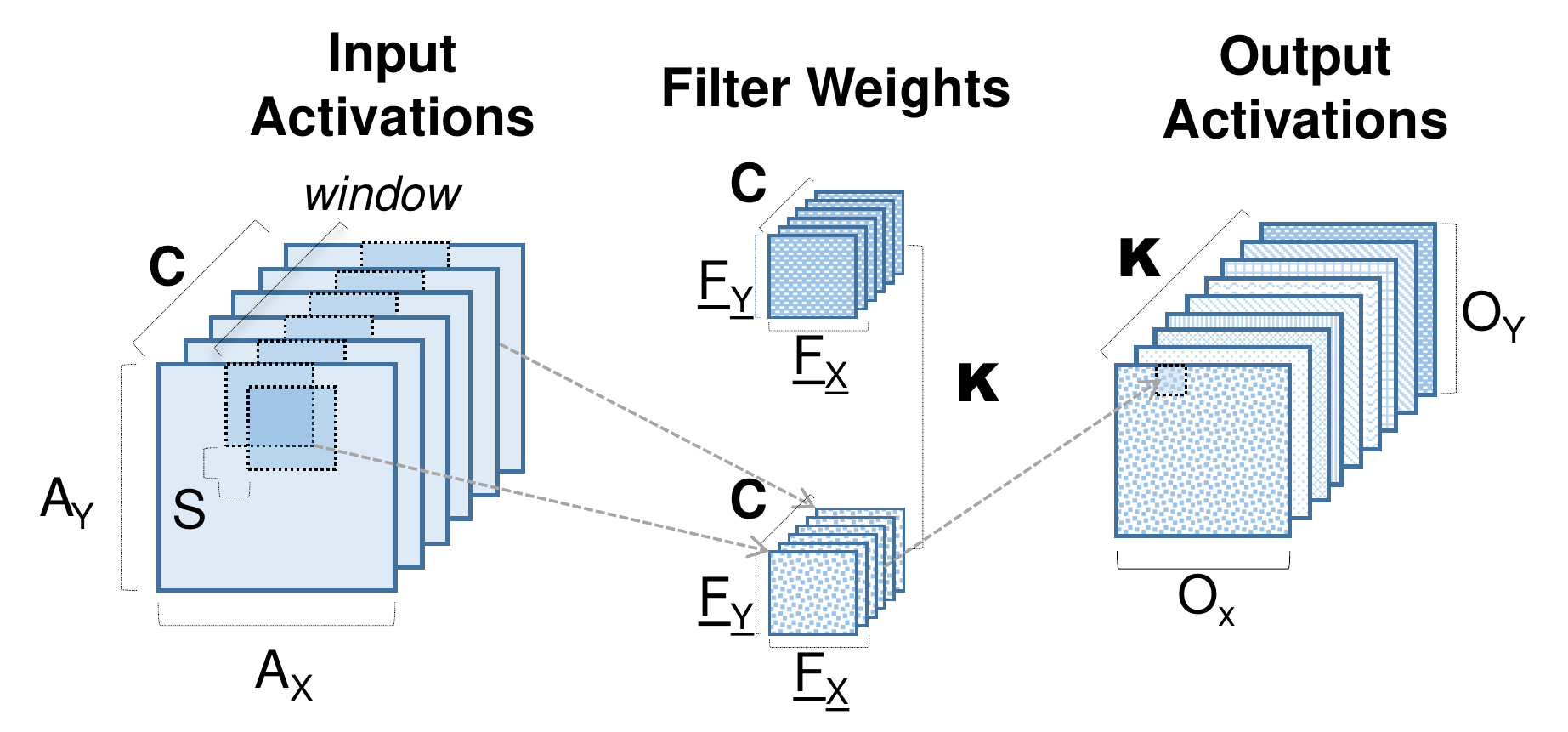}
\caption{Convolutional Layer.}
\label{fig:CNN}
\end{figure}

A CNN comprises a chain of layers. While there are several layer types, convolutional layers (CVLs) dominate execution time for many image related applications~\cite{diannao}.
Figure~\ref{fig:CNN} illustrates a convolutional layer. An $A_X \times A_Y \times C$  \textit{activation} array $A$ is convolved with $K$ $F_X \times F_Y \times C$ \textit{weight} \textit{filters} (denoted $F^0\cdots F^{K-1}$) producing an output $O_X \times O_Y \times K$ activation array $O$. Each output activation $o(x,y,k)$ is the dot product of filter $F^k$ with a \textit{window}, which is an $F_X \times F_Y \times C$ subarray of the input activation array. Windows are evenly spaced using a stride $S$ resulting in a grid of $\lceil A_X/S\rceil \times \lceil A_Y/S\rceil= O_X \times O_Y$. Each output passes through an activation function, typically, the Rectified Linear Unit (ReLU). A typical layer requires 100s to 1,000s dot product calculations, each of 100s to 1,000s of input weight and activation pairs.

\begin{figure}
\begin{verbatim}
for wx = 0 to Ax-Fx step S
  for wy = 0 to Ay-Fy step S
    for c = 0 to C-1
      for fx = 0 to Fx-1
        for fy = 0 to Fy-1
          for k = 0 to K-1
              ax = wx + fx
              ay = wy + fy
              ox = wx / S
              oy = wy / S
              O[ox, oy, c] +=
               F[k][fx, fy, c] * A[ax,ay,c]
\end{verbatim}
\caption{Convolutional Layer Calculation.}
\label{fig:cnn:code}
\end{figure}

Figure~\ref{fig:cnn:code} shows an implementation of a CVL as a 6-nested loop -- batching, that is the processing of multiple inputs, adds another loop. Any permutation of the loops is possible. Tiling can improve locality which is especially useful when only a limited amount of on-chip storage is available~\cite{EyerissISCA2016}. Finally, since the multiplications are independent, there are many different ways in which these loops can be executed concurrently.

A fully-connected layer can be implemented as a convolutional layer with a single window and where the filters and the input are of the same dimensions.

\begin{figure}
\includegraphics[width=0.40\textwidth]{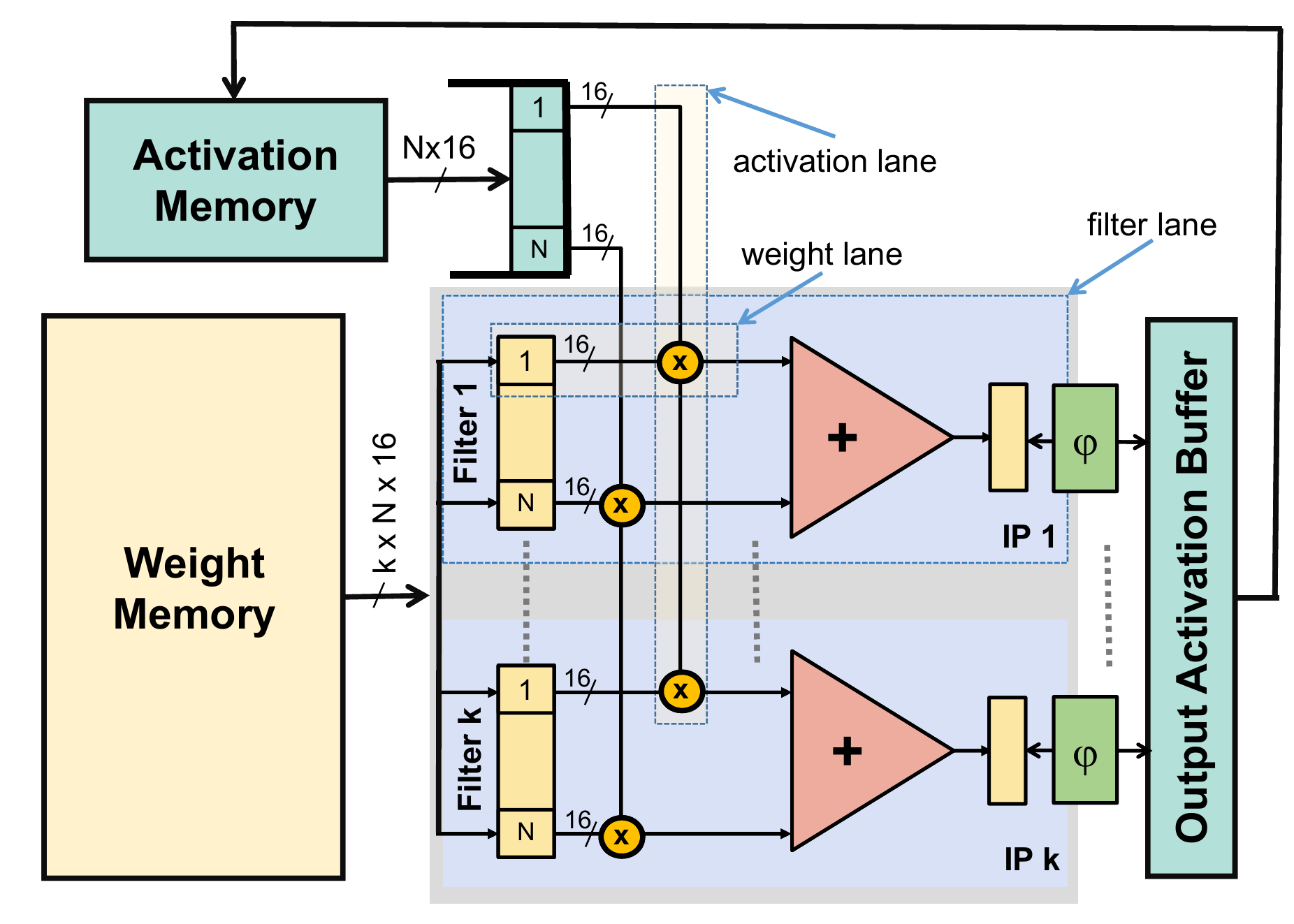}
\caption{Dense NN Accelerator Tile}
\label{fig:base}
\end{figure}

\subsection{Sparsity}
\label{sec:sparsity}
Figure~\ref{fig:cnn:code} assumes a \textit{Dense} CNN since it processes all weights and activations regardless of their value. In practice some weights are zero and some activations are zero (or close enough to zero that can be treated as such), and hence are ineffectual~\cite{isscc_2016_chen_eyeriss,han_eie:isca_2016, albericio:cnvlutin,Reagen2016,cambricon:2016}. Weight pruning, which often requires retraining the CNN, can further increase weight sparsity resulting into a \textit{Sparse} CNN~\cite{han_deep_2015-1,DynamiCNetworkSurgery}.

\begin{figure}
\centering
\subfloat[][Dense Filter]{
\centering
\includegraphics[width=0.23\textwidth]{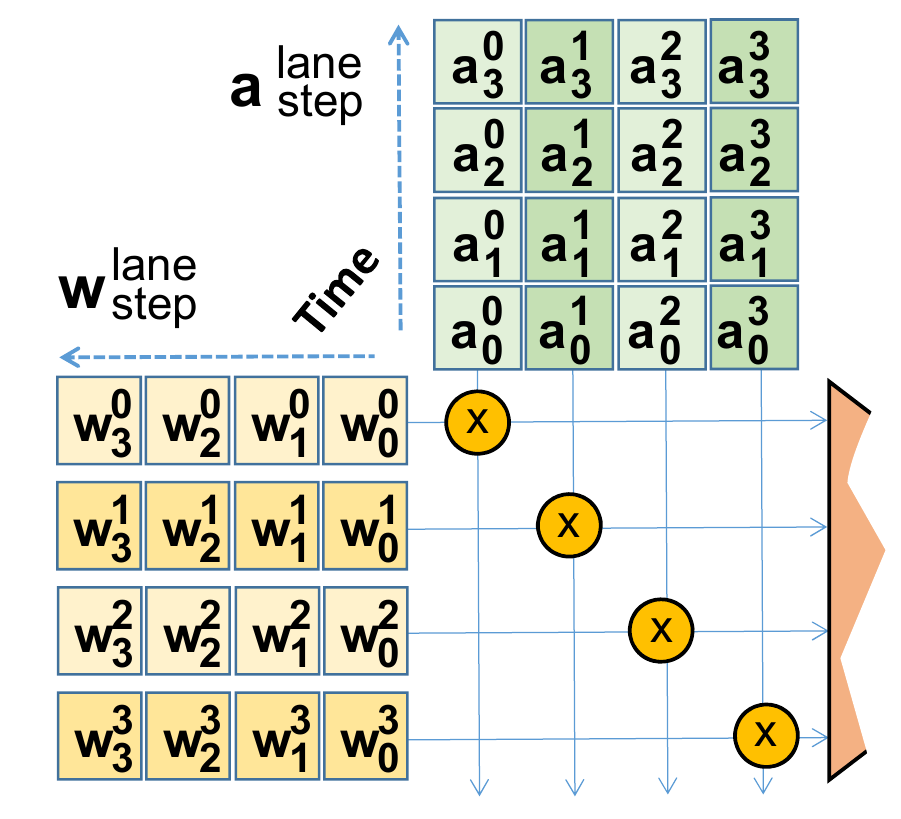}
\label{fig:sdd}
}
\subfloat[][Sparse Filter]{
\centering
\includegraphics[width=0.23\textwidth]{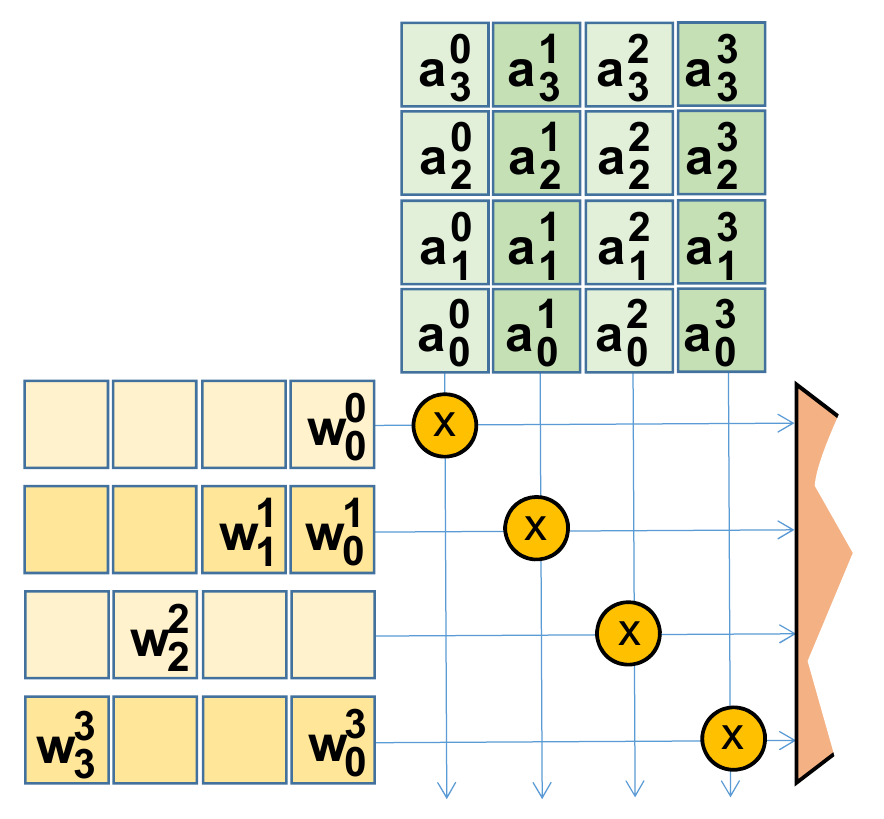}
\label{fig:sds}
}
\caption{A Dense Accelerator with 4 multipliers takes 4 cycles to process a 16-term dot-product regardless of whether it is (a) from a Dense CNN, or (b) from a Sparse CNN.}
\label{fig:sd}
\end{figure}

\subsection{Dense CNN Accelerator}
\label{sec:base}

Figure~\ref{fig:base} shows a data-parallel accelerator for dense CNNs that produces $k$ partial output activations per cycle. It comprises $k$ inner product units (IPUs) operating in parallel over the same set of $N$ activations. Each input activation is multiplied with $k$ weights, one per filter as follows: Each IPU accepts a vector of $N$ weights per cycle, one per input activation, calculates $N$ products, reduces them via an adder tree, and accumulates the result into an output register. Once a full window has been processed, usually over multiple cycles, the output register contains the corresponding output activation.

A \textit{Weight Memory} (WM) and an \textit{Activation Memory} (AM) respectively provide the weights and the activations. For the purposes of this work we assume a memory hierarchy similar to DaDianNao~\cite{DaDiannao}: 1)~AM and WM are large enough to hold a full layer at a time similar to DaDianNao~\cite{DaDiannao} and the TPU~\cite{TPUISCA17}, 2)~WM can supply $N\times k$ weights per cycle via a single, wide read port, 3)~AM can supply $N$ activations per cycle via a wide read port, 4)~weight and activation buffers hide the latency of AM and WM, and 5)~an output activation buffer collects the results prior to writing them back to AM. All designs use a 16-bit fixed-point format to represent activations and weights, a common design choice for most inference accelerators. For clarity, we assume that when multiple tiles exist, they are all connected to the same AM which broadcasts a block of $N$ activations per cycle to all tiles. We will use the acronym \textit{\BASE} to refer to configurations of this accelerator.

Figure~\ref{fig:sdd} shows a \textit{schedule} of how an example \BASE with $k=1$ and $N=4$ would process 4 sets of 4 activation and weight pairs. Activations and weights are respectively denoted with $a^{lane}_{step}$ and $w^{lane}_{step}$ where $lane$ designate the activation column and the weight row they appear at, and $step$ is the order in time in which they get multiplied. In the text, we will use the notation $a[lane,step]$ and $w[lane,step]$. Assuming that each step requires a single cycle, \BASE would process the 16 products in 4 cycles. Figure~\ref{fig:sds} shows how \BASE would process a sparse CNN where some of the weights are zero (shown as empty boxes in the figure). Since \BASE has no provision for skipping ineffectual weights, it still takes 4 cycles to process the input. However, only 6 of the 16 products involve an effectual weight. If it were possible to freely schedule those 6 products over the four multipliers 2 cycles would be sufficient.

The first goal of \OUR is to deliver a sufficient portion of the benefits possible from eliminating products with ineffectual weights while avoiding the complexities of an unrestricted schedule of weights and activations.

\section{Exploiting Weight Sparsity}
\label{sec:ews}

This section explains how \OURS eliminates enough ineffectual weights. Ineffectual weight elimination in \OURS is scheduled \textit{statically} by \textit{promoting} effectual weights in time placing them in the position originally occupied by an ineffectual weight. A software scheduling pass rearranges the weights once prior to deployment so that they appear at the appropriate weight lane and at the right step when \OURS fetches them at runtime. As a result, a \OURS tile can access all $k\times N$ weights it needs per step with a single wide access to WM. Each effectual weight carries with it a narrow piece of metadata that identifies its position in the original dense schedule so that at runtime \OURS can match it with the appropriate activation.

\OURS strikes a balance between weight scheduling flexibility and energy efficiency. This it achieves by allowing schedules where only two  \textit{intra-filter} weights \textit{``movements''} are legal: lookahead and lookaside. With \textit{lookahead} an effectual weight $w[lane, step]$ may advance to replace any ineffectual weight up to $w[lane, step-h]$ while staying at the same weight lane. With \text{lookaside} a weight can advance ``sideways'' by $d$ lanes and by one step in time. That is, a weight $w[lane,step]$ may replace any weight at $w[(lane+d)\ MOD\ (N - 1), step-1]$. Sections~\ref{sec:ws:lookahead} and~\ref{sec:ws:lookaside} explain lookahead and lookaside through an example and Section~\ref{sec:ws:impl} presents their implementation.
While this section restricts lookaside to one step ahead of time, this is done for ease of explanation.  Section~\ref{sec:e:la:other} considers other lookaside schemes which may advance weights up to $h$ steps ahead of time that prove advantageous.

For clarity, the rest of this section assumes bit-parallel multipliers and 16 input/communication wires per activation. The designs of Section~\ref{sec:ea} however will use only few wires per activation, 1 or 4 depending on the design. Furthermore, the description uses $h$ and $d$ to refer to the options available for lookahead and lookaside. In practice $2$ and $5$ respectively prove sufficiency.

\begin{figure*}
\centering
\subfloat[][Cycle 0]{
\centering
\includegraphics[width=0.3\textwidth]{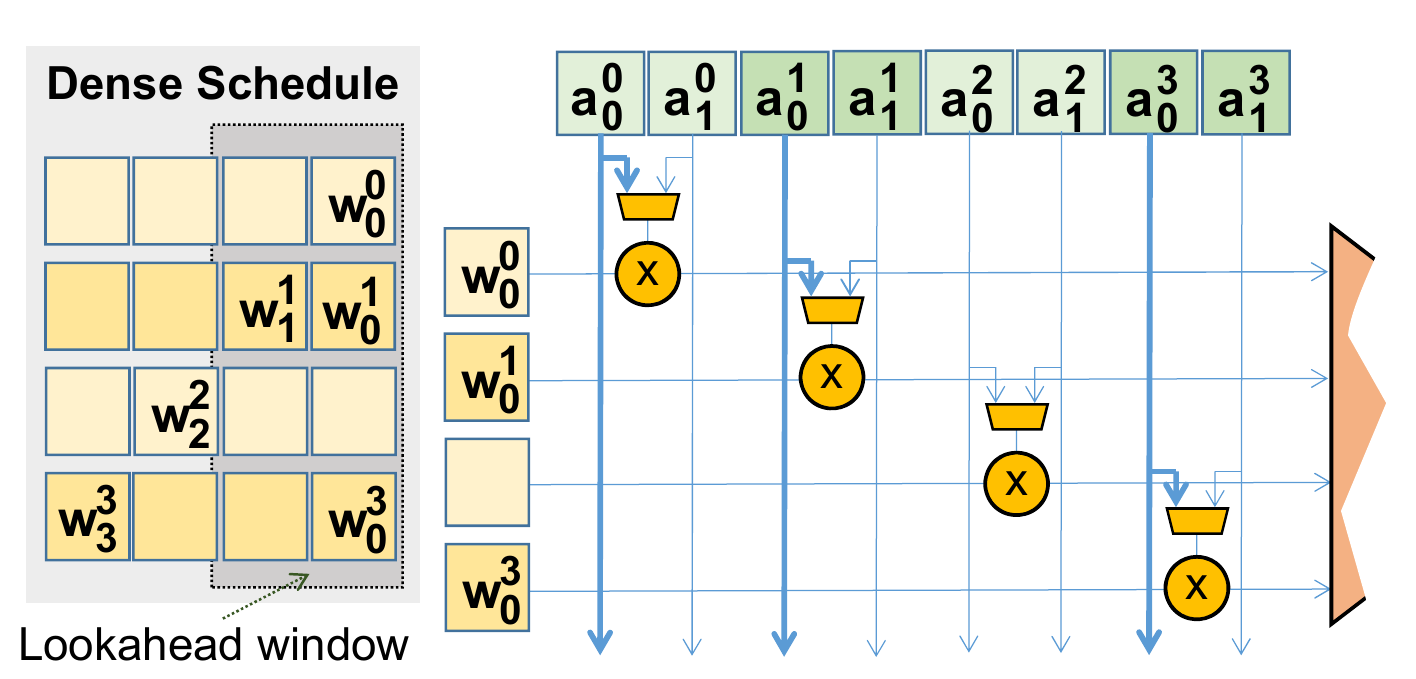}
\label{fig:sslh0}
}
\subfloat[][Cycle 1]{
\centering
\includegraphics[width=0.3\textwidth]{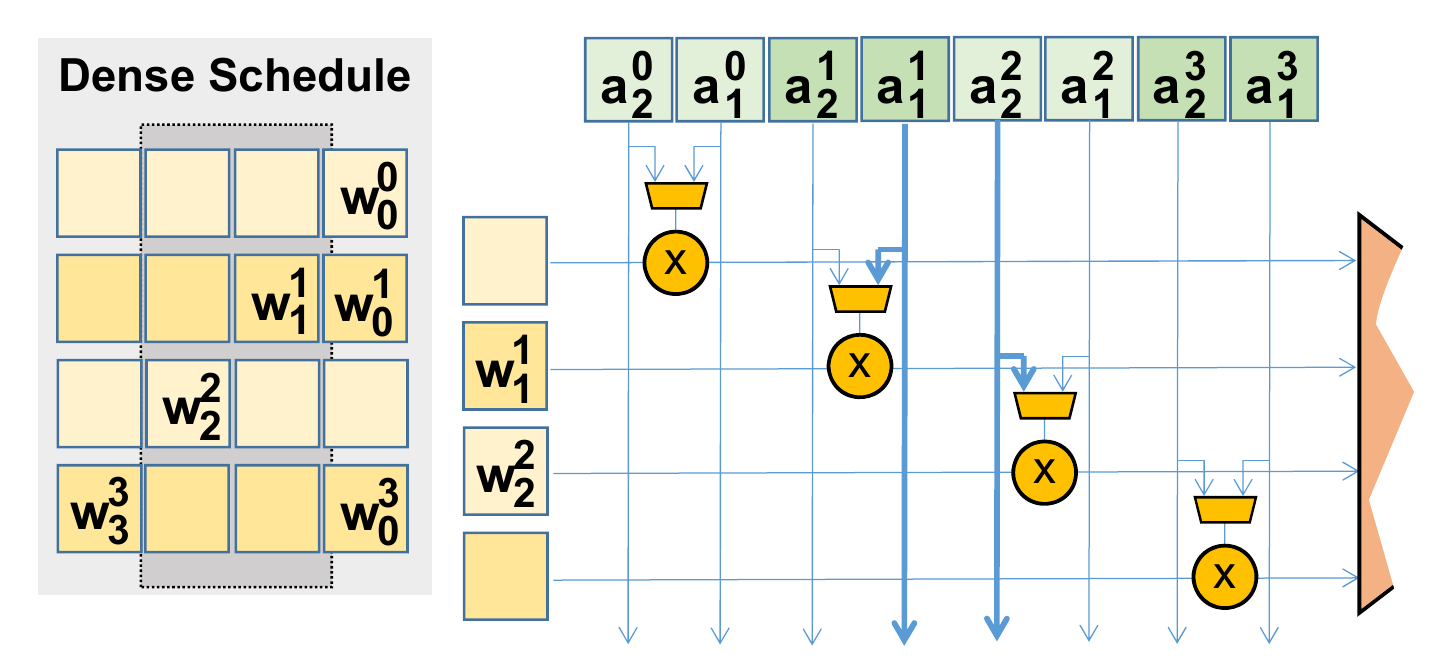}
\label{fig:sslh1}
}
\subfloat[][Cycle 2]{
\centering
\includegraphics[width=0.3\textwidth]{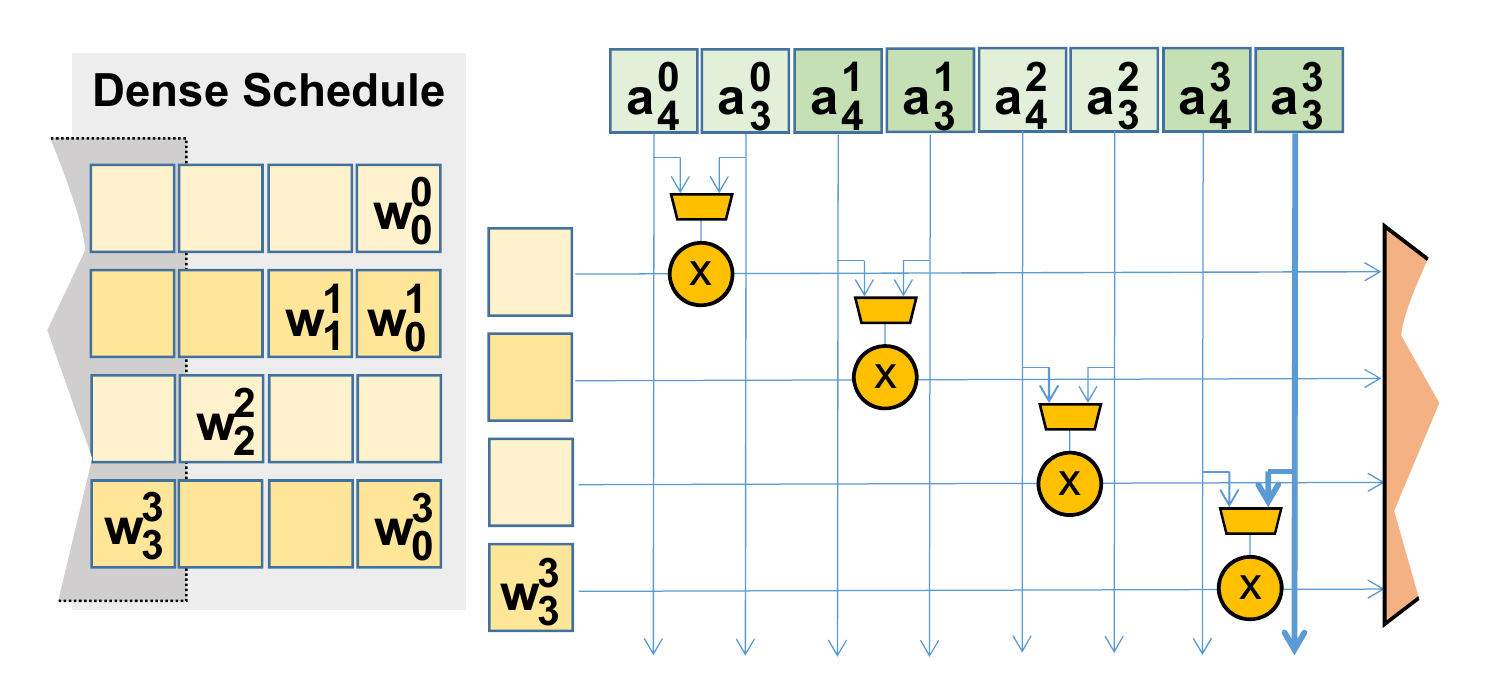}
\label{fig:sslh2}
}
\caption{\OURS Accelerator with Lookahead of 1 processes the sparse NN of Fig.~\ref{fig:sds} in 3 cycles.}
\label{fig:sslh}
\end{figure*}

\begin{figure*}
\centering
\subfloat[][Cycle 0]{
\centering
\includegraphics[width=0.3\textwidth]{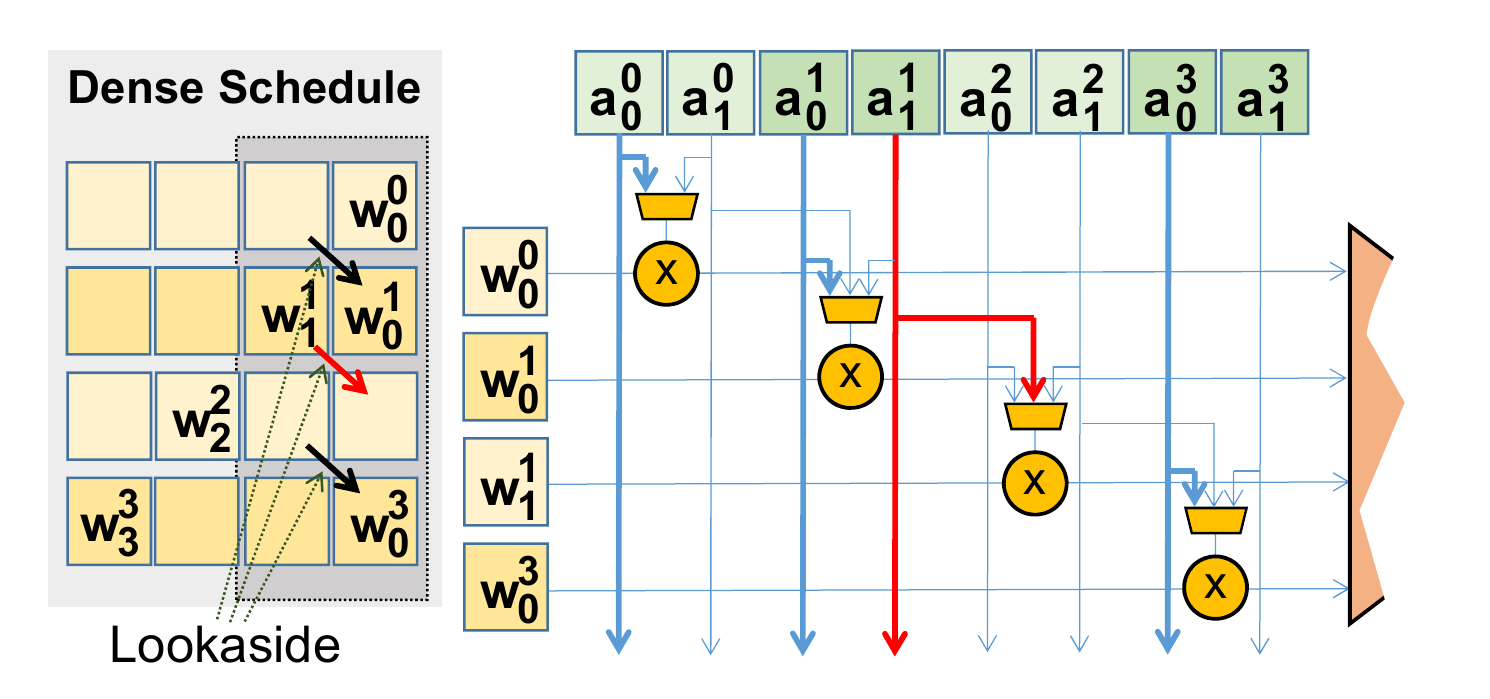}
\label{fig:ssla0}
}
\subfloat[][Cycle 1]{
\centering
\includegraphics[width=0.3\textwidth]{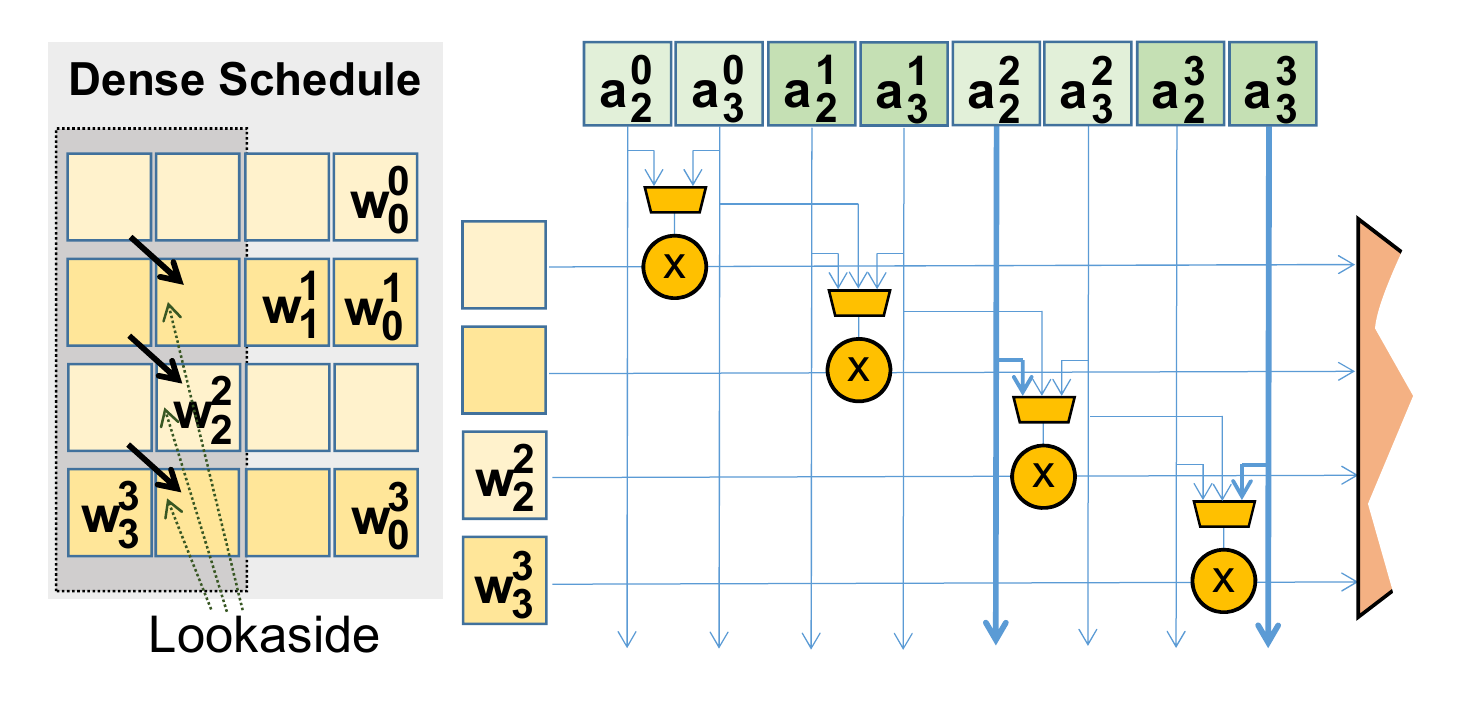}
\label{fig:ssla1}
}
\caption{\OURS with Lookahead of 1 and Lookaside of 1 processes the sparse NN of Fig.~\ref{fig:sds} in 2 cycles, the minimum possible.}
\label{fig:ssla}
\end{figure*}

\subsection{Weight Lookahead}
\label{sec:ws:lookahead}
Figure~\ref{fig:sslh} shows an example of lookahead for the sparse filter of Figure~\ref{fig:sds}. The \BASE schedule from Figure~\ref{fig:sds} is shown for reference, and parts~(a) through~(c) illustrate how lookahead $h=1$ reduces execution time to 3 cycles. Like \BASE, the example \OURS configuration can process 4 products per cycle.  Conceptually, lookahead amounts to establishing a sliding window of $h+1$ within which weights can be promoted over ineffectual weights that appear on the same lane. In Figure~\ref{fig:sslh0} and cycle 0, lookahead fails to utilize weight lane 2 since weight $w[2,2]$ is at lookahead distance 2. In Figure~\ref{fig:sslh1} and cycle 1, lookahead promotes $w[2,2]$ to replace $w[2,1]$. However, $w[3,3]$ is out of reach as lane 1 is now processing $w[1,1]$ limiting lookahead to weights that appear up to step 2 in the dense schedule. As there are no longer any weights to process in step 2, the lookahead window now progresses two steps such that, finally, in Figure~\ref{fig:sslh2} and cycle 2, $w[3,3]$ is processed.

Since weights are statically promoted along their respective lane, \OURS must pair them with the corresponding activation at runtime. As the figure shows, to achieve this pairing \OURS requires that all activations for the full lookahead window be available. For each weight lane, there are now 2 activation lanes corresponding to time steps $t$ and $t+1$. \OURS selects the appropriate activation via a per weight lane \textit{2-to-1} multiplexer. The control signal for the multiplexer is determined statically when the schedule is created and is stored along with the weight. In general, for a lookahead $h$ and per weight lane, \OURS uses $h$ extra activation lanes and an \textit{(h+1)-to-1} multiplexer to select the appropriate activation. Section~\ref{sec:ws:impl} explains how supporting a wider group of activation leads to a practical and low cost implementation. For time being suffices to recall that the activation lanes are shared among $k$ filters per tile and thus their cost is amortized over multiple weight lanes. Moreover, Section~\ref{sec:e} shows that most of the benefits possible with lookahead can be had with $h\leq 2$.

\subsection{Weight Lookaside}
\label{sec:ws:lookaside}
While lookahead reduces execution time, it does suffer from lane imbalance as the lane with the more effectual weights will determine the number of steps needed to process a window. Lookaside introduces additional flexibility by allowing a weight lane to ``steal'' work from another lane from the same filter.  Recall that all weight lanes per filter feed into the same adder tree and thus moving weights around lanes does not affect the outcome. Figure~\ref{fig:ssla} shows how our example \OURS works with lookaside $d=1$. Here an otherwise idle lane can ``steal'' an effectual weight from its immediately neighboring lane and from the next time step. In Figure~\ref{fig:ssla0} and at cycle 0, lane 2 ``steals'' $w[1,1]$ from lane 1 and avoids staying idle while also allowing the lookahead window to progress by two steps leading to cycle 1 in Figure~\ref{fig:ssla1}. As a result, in cycle 1 and through lookahead, lane 3 can process $w[3,3]$ at the same time as lane 2 is processing $w[2,2]$. In total, the example \OURS with $h=d=1$ takes 2 cycles to process the input, which is the minimum number of steps possible.

Lookaside requires no additional activation lanes when lookahead is at least 1. It just requires an activation multiplexer with more inputs. In general, it needs \textit{(h+d+1)-to-1} multiplexers for lookaside $h$ and lookahead $d$. As Section~\ref{sec:ws:impl} explains, the data input connections for these multiplexers are statically determined and regular. As with lookahead, the control signal for the multiplexer is determined statically and stored along with the weight and it requires $lg(h+d+1)$ bits. The result section shows that $h+d+1=8$ is sufficient. Adding the ability to ``steal'' weights from steps beyond $t+1$ is possible and it would introduce more scheduling flexibility albeit at an increased interconnect cost.

\subsection{Implementation}
\label{sec:ws:impl}

\begin{figure}
\centering
\subfloat[][Tile.]{
\centering
\includegraphics[width=0.43\textwidth]{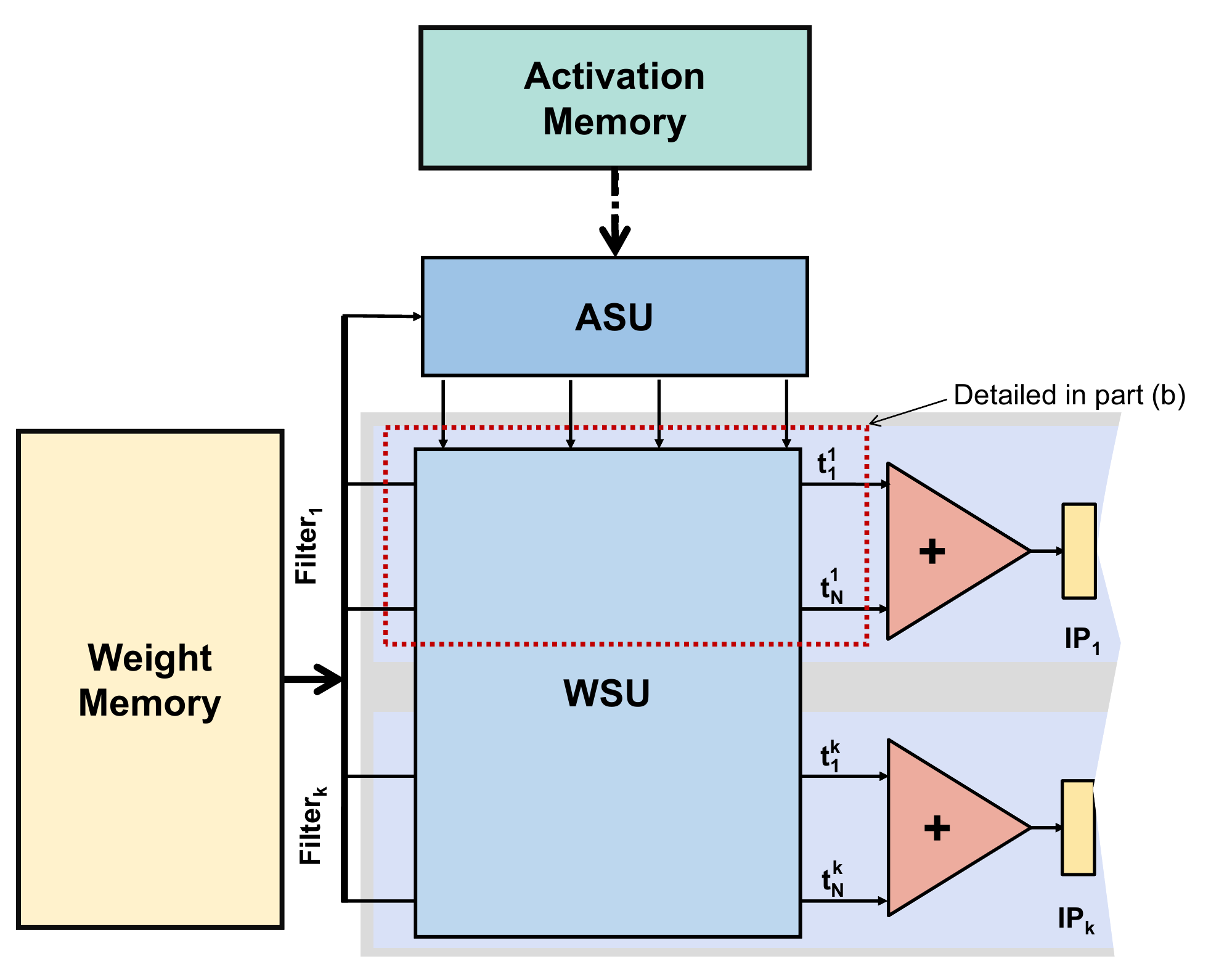}
\label{fig:wsi:tile}
}\\ \vspace{-0.3cm}
\subfloat[][WSU: Weight Skipping Unit Slice.]{
\centering
\includegraphics[width=0.5\textwidth]{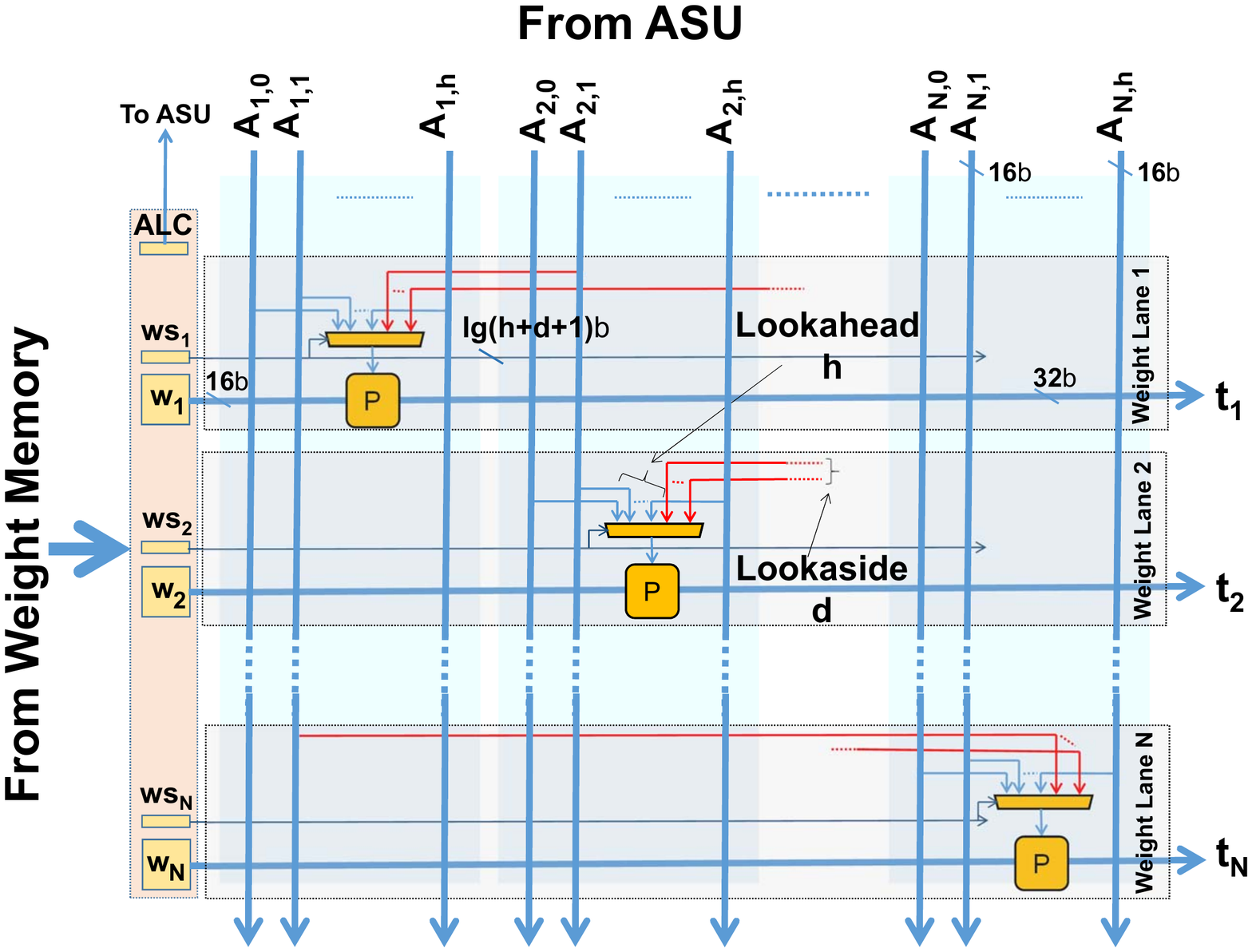}
\label{fig:wsi:wsu}
}\\ \vspace{-0.3cm}
\subfloat[][ASU: Activation Selection Unit.]{
\centering
\includegraphics[width=0.5\textwidth]{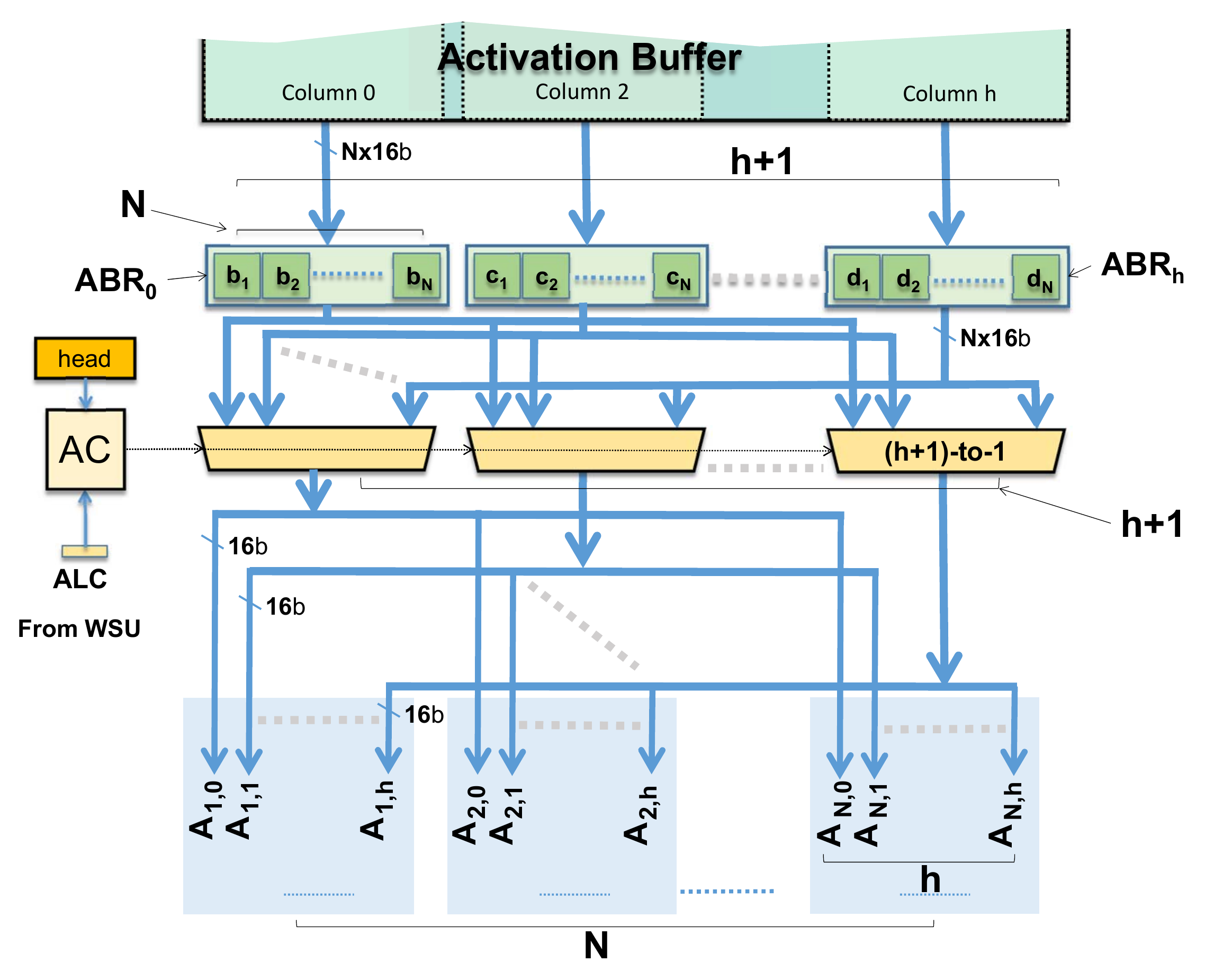}
\label{fig:wsi:asu}
}\\
\caption{Weight Skipping Accelerator Tile Architecture. The figure assumes 16b activations. Section~\ref{sec:ea} will reduce the number of wires needed per activation to 1 or 5.}
\label{fig:wsi}
\end{figure}

Figure~\ref{fig:wsi} shows how \OURS implements weight skipping. Figure~\ref{fig:wsi:tile} shows a \OURS tile that processes $N$ products per filter in parallel for $k$ filters. An \textit{Activation Select Unit} (ASU) buffers activations as they are provided by the AM and rearranges them so that the \textit{Weight Skipping Unit} (WSU) can straightforwardly select the appropriate activations.

\noindent\textbf{Weight Skipping Unit:}
Figure~\ref{fig:wsi:wsu} shows a WSU slice. There is one slice per filter for a total of $k$ slices per tile. The WSU reads via single WM port a column of prescheduled weights along with their multiplexer select metadata.  In total, the WSU reads $N \times k$ weight and metadata pairs plus an activation lane control (ALC) field per access. The next section explains the use of the ALC. In this column there are $N$ weights per WSU slice and all weights are processed in one step.

Focusing on a single filter slice and as the figure shows, each of the $w_1$ through $w_N$ weights maps onto a separate weight lane where it feeds one of the inputs of a multiplier P. An \textit{(h+d+1)-to-1} multiplexer selects the second input to the multiplier. The multiplexer control signal comes from the corresponding \textit{weight select} metadata ($ws_1$ through $ws_N$ on the figure) that the WSU reads from WM. The first multiplexer input allows a weight to stay at its original dense schedule position, another $h$ inputs implement lookahead and the final $d$ inputs implement lookaside.

For each weight $w_i$ there are $h+1$ activations, $A_{i,0}$ through $A_{i,h}$, that correspond to a lookahead window of $h$ activations. For example, for $w_1$, $A_{1,2}$ is the activation that is at lookahead 2, whereas for $w_N$, $A_{N,h}$ is the activation at lookahead $h$. As the next section explains, the ASU ensures that the physical order of the activations coincides with their logical lookahead order. This allows WSU to implement lookahead and lookaside by statically assigning $A_{i,j}$ signals to multiplexer inputs. For example, the lookaside 1 connection for $w_2$ is to $A_{3,1}$ and its lookahead 2 connection is to $A_{2,2}$. All WSU slices share the same $(h+1) \times N$ activations.

In the figure, the WSU slice produces $N$ $16b\times 16b$ products per cycle ($t_1$ through $t_N$). Those feed an adder tree whose output accumulates into an output activation over multiple cycles. Section~\ref{sec:ea} replaces the multipliers P with AND gates or shifters to take advantage of activation properties.

\noindent\textbf{Activation Select unit:}
Figure~\ref{fig:wsi:asu} shows the ASU whose purpose is to generate the $A_{lane,lookahead}$ signals the WSU needs. Its goal is to ensure that each $A_{lane,lookahead}$ contains the input activation needed by the corresponding weight $lane$ and a step distance $lookahead$.  It contains $h+1$ \textit{Activation Block Registers} (ABRs) each holding $N$ input activations. Each ABR contains the $N$ activations needed by all weight lanes at some specific lookahead distance $l=0$ to $h$. The ABRs operate logically as a circular queue with the $head$ register pointing to the ABR holding the activations at $lookahead=0$. An array of $h+1$ \textit{(h+1)-to-1} multiplexers shuffle the ABR outputs to the appropriate order generating the $A_{lane,lookahead}$ signals which are distributed along the weight columns as the figure shows.
The ALC metadata the WSU reads from WM along with each $N\times k$ weight column is used to advance the $head$ register and implements the sliding lookahead window.

An \textit{Activation Buffer} (AB) buffers activations as they are read from AM. The AB has $h+1$ banks, each connected to one ABR via a dedicated single read port. This way, any number of ABRs can be updated per cycle concurrently effectively advancing the lookahead window as instructed by the ALC metadata. This arrangement enables \OURS to also skip over columns comprising ineffectual weights only.

\section{Exploiting Activations}
\label{sec:ea}
While weight skipping exploits weight sparsity it does not exploit any of the potentially valuable properties of the activations. We present two \OURS variants, \OURPRA and \OURSTR that deliver different area, performance, and energy efficiency trade offs. \OURPRA exploits the effectual bit content of activations and given enough resources outperforms \OURSTR. \OURSTR exploits fine-grain dynamic activation precision variability and requires fewer resources than \OURPRA. Compared to \BASE both designs deliver benefits for \textit{all} activations, ineffectual or not, they deliver benefits  for both \textit{sparse} and \textit{dense} CNNs, and can execute fully-connected layers without modification.  Section~\ref{sec:apra} discusses \OURPRA and Section~\ref{sec:astr} discusses \OURSTR. Neither design attempts to eliminate ineffectual activations. However, as Section~\ref{sec:ia:whynot} explains, both deliver some of the benefits possible from ineffectual activations while at the same time benefiting also from effectual activations.

\subsection{\OURPRAcore: Effectual Bit Content}
\label{sec:apra}

\OURPRA is motivated by the observation that in dense CNNs on average more than 90\% of the activation \textit{bits} are zero and thus are ineffectual during multiplication~\cite{Pragmatic}. Even if zero activations could be eliminated, the fraction of ineffectual activation bits remains higher than 75\%. Section~\ref{sec:motivate} showed that this lopsided distribution of effectual vs. ineffectual activation bits persists in the sparse NNs studied here.

\OURPRA's goal is to exploit this abundance of ineffectual activation bits to complement the benefits delivered by eliminating ineffectual weights. For this purpose, \OURPRA aims to process \textit{only} the effectual bits of activations bit-serially over time. For example, ideally, \OURPRA will process the activation value \texttt{\{0000 0000 1000 1111b\}} over 3 cycles respectively multiplying the corresponding weight by the following signed powers of two: \texttt{\{+2\textsuperscript{7}, +2\textsuperscript{4}, -2\textsuperscript{0}\}}. These powers are the Booth-encoded representation of the activation value.  In general, this approach results in execution time that is proportional to the number of effectual activation bits.  However, since now activations are processed bit-serially, overall throughput will be lower if \OURPRA processes only $N\times N$ weight and activation pairs per cycle.  \OURPRA compensates for this loss of throughput by processing 16 activation windows in parallel thus always matching or exceeding \BASE's throughput. As a result, the same weight can be reused over the 16 windows and the WM interface remains as-is. However, all lanes that feed a common adder tree are best to remain synchronized across activation groups, that is, all have to wait for the one processing the activation with the most effectual bits to finish before proceeding with the next group of activation values.

\subsubsection{\OURPRAcore\ Implementation}
\OURPRA adapts the \textit{Pragmatic} accelerator (\PRAS) design for processing activations~\cite{Pragmatic}. \PRAS targets dense CNNs and exploits ineffectual activation bits to deliver execution time that is proportional to the effectual activation bit content. To do so, \PRAS processes activations bit-serially, one effectual bit at a time. A per tile unit converts the activations into a stream of effectual powers of two, or \textit{oneffsets} after applying a modified Booth Encoding. Since \PRAS multiplies a weight with a power of two each cycle, a shifter is sufficient instead. The oneffsets sign is used to add or subtract the shifted weight via the adder tree. To guarantee that \PRAS always matches or exceeds the throughput of an equivalent bit-parallel design, \PRAS processes concurrently 16 activation windows. This allows \PRA to reuse the same weight across 16 IP units. Albericio \textit{et al.,} detail the \PRAS design~\cite{Pragmatic}.

\begin{figure}
\centering
\includegraphics[width=0.40\textwidth]{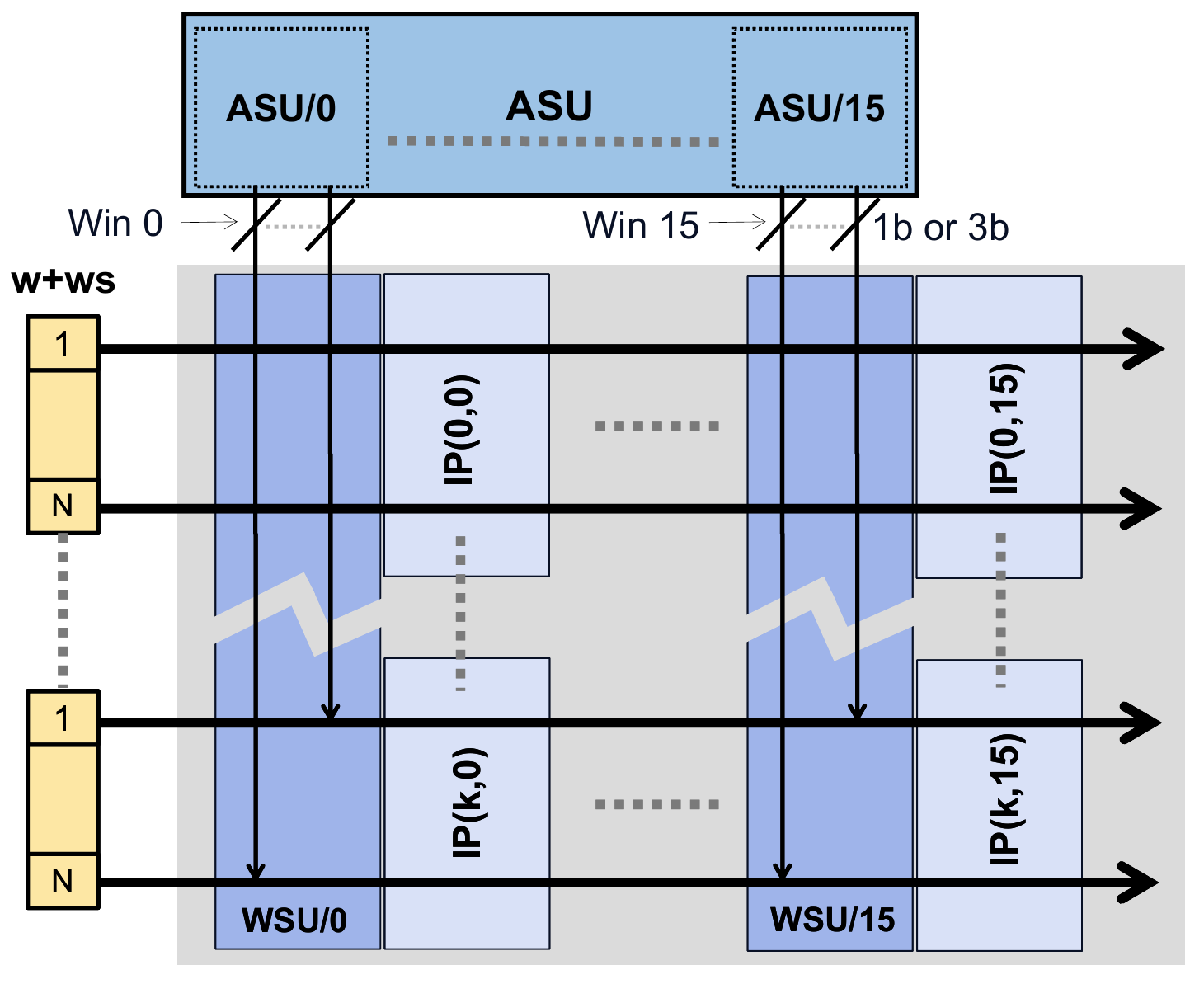}
\caption{Adding the capability to skip ineffectual activation bits.}
\label{fig:wsi:pra}
\end{figure}

Figure~\ref{fig:wsi:pra} shows a \OURPRA tile. The $k$ IP units have been expanded into a grid of $16\times k$ simpler IP units. The original WSU is sliced in 16 columns, WSU/0 through WSU/15, one per column of $k$ IPs. Each of those 16 columns corresponds to a different window. Each IP has a 16-input adder tree and instead of $N$ multipliers it has $N$ shifters. Each of these shift the 16b weight input as directed by the activation oneffset input. All IPs along the same row share the same $w$ and $ws$ signals and they all perform exactly the same lookahead and lookaside activation selections. Whereas in the design of Figure~\ref{fig:wsi} the per weight lane multiplexers selected a 16b activation, here, the per weight lane and per IP multiplexers select a 3b activation oneffset. These oneffsets encode a shift by up to 3 positions plus a sign. For each column, a corresponding ASU slice provides as before data for $N$ activation groups, one per weight lane, each containing data for $h$ activations to support lookahead. However, unlike Figure~\ref{fig:wsi} instead of 16b activations, the ASU provides 3b oneffsets. Since all WSU columns execute the same weight schedule, all 16 ASU slices access the activation buffer in tandem and share the same activation selection logic and signals.

\subsection{\OURSTRcore: Precision}
\label{sec:astr}

An alternative to exploiting the effectual bit content of activations is to exploit their precision requirements. The precision activations need varies across networks and across layers and can be determined through profiling, e.g.,~\cite{judd:reduced}. \OURSTR aims to exploit this precision variability to reduce execution time proportionally. Compared to the baseline precision of 16b, \OURSTR would ideally reduce execution time by $16/p$ where $p$ is the precision activations use. For this purpose, \OURSTR adopts the \textit{Stripes} (\STR) accelerator tile design~\cite{Stripes-MICRO}. \STR processes activations bit-serially and thus takes $p$ cycles to process an activation represented in $p$ bits. As with \PRA, to compensate for the loss in computation bandwidth compared to a bit-parallel design, \STR processes 16 windows in parallel. This way it guarantees that its execution time will be at most the same as that of an equivalent data parallel design. Instead of multipliers, \STR uses AND gates. The implementation of \OURSTR is virtually identical at the block level to that of \OURPRA which was shown in Figure~\ref{fig:wsi:pra}. The primary difference is that the ASU now sends activations a single bit at a time instead of a single oneffset at a time, and does not need to encode activations as oneffsets. The overall cost is lower. Fewer wires are required per activation, there are no shifters, and the input width of the adder tree is 16b.

\subsubsection{Dynamic Precision Reduction}
While \STR used profile-derived precision requirements, Lascorz \textit{et al.,} observed that a profile-derived precision for a layer is bound to be pessimistic for two reasons: 1)~the precision must accommodate any possible input, and 2)~the precision must accommodate all activations for the layer~\cite{dynamicstripes}. However, in practice only a limited set of activations for one specific input will be processed concurrently at runtime. Moreover, given that most activations tend to be near zero, this approach significantly reduces the precision needed per group of concurrently processed activations. The precision needed for each activation group is detected when the precisions are read from the AM and communicated along with activation values. Both \OURSTR and \OURPRA incorporate this technique. For \OURSTR dynamic precision reduction reduces execution time while for both \OURSTR and \OURPRA it reduces the number of bits that needs to be sent after reading the activations from AM. Recall that \OURPRA generates oneffsets locally at each tile.

\subsection{Interaction with Weight Elimination}
While increasing lookahead may eliminate more weights, it will also decrease the effectiveness of the activation optimizations.  Specifically, in both \STR and \PRA a group of concurrently processed activations has to wait for the slowest activation to process before advancing for the next group. For example, in \PRA it is the activation with the highest number of powers of two that determines how many cycles would be required for the whole group. As the degree of lookahead increases, \OURPRA and \OURSTR have to consider all activations within the lookahead window. The wider the lookahead window the higher the impact of such ``bottleneck'' activations. Lookaside has no further effect as it uses the activations at a lookahead distance of 1 which are included in the synchronization group when lookahead is at least 1.

\subsection{Ineffectual Activations?}
\label{sec:ia:whynot}
Nearly half of the activations in CNNs tend to be ineffectual~\cite{albericio:cnvlutin,han_eie:isca_2016,EyerissISCA2016,SCNN} yet \OURS does not attack ineffectual activations head on. However, by attacking dynamic precision variability or ineffectual bit content, \OURS delivers benefits for both ineffectual and effectual activations alike. Let us consider a hypothetical bit-parallel \textit{DSPARSE} accelerator that can skip all ineffectual activations. Clearly for the effectual activations, \OURS will be at an advantage since \textit{DSPARSE} does not target them. However, anytime \OURS does not completely skip an ineffectual activation this represents an opportunity loss compared to \textit{DSPARSE}. The key insight is that on average, this opportunity loss will be a lot less than ``one unit of work'' per ineffectual activation. Consider for example, the extreme, yet illustrative case where all activations that are processed as a group by \OURPRA happen to be zero. \OURPRA will process them in a single cycle which represents an opportunity loss of only $1/16$ vs. \textit{DSPARSE} since \OURPRA processes each activation bit-serially. In general, when \OURPRA processes an ineffectual activation over $p$ cycles, the opportunity loss is $p/16$. Given that on average less than 10\% of the bits are effectual, the opportunity loss of not completely skipping ineffectual activations is expected to be low. A similar reasoning applies to \OURSTR as well.

In either case, ineffectual activations, dynamic precision variability and ineffectual activation bits are consequences of the distribution of activation values in CNNs: often most activations cluster near zero and a few spike with values far away from it. For image classification CNNs, on average about 45\% of activations were found to be zero even after reducing their precision per layer~\cite{albericio:cnvlutin}. In contrast, on average more than 90\% of the activation bits are found to be zero suggesting that the potential for performance improvement is much higher if we target ineffectual bit content. Dynamic precision reduction acts as a lower-cost proxy for ineffectual bit content.

\section{Multi-Tile Architecture}
\label{sec:full}

\begin{figure}
\centering
\includegraphics[width=0.35\textwidth]{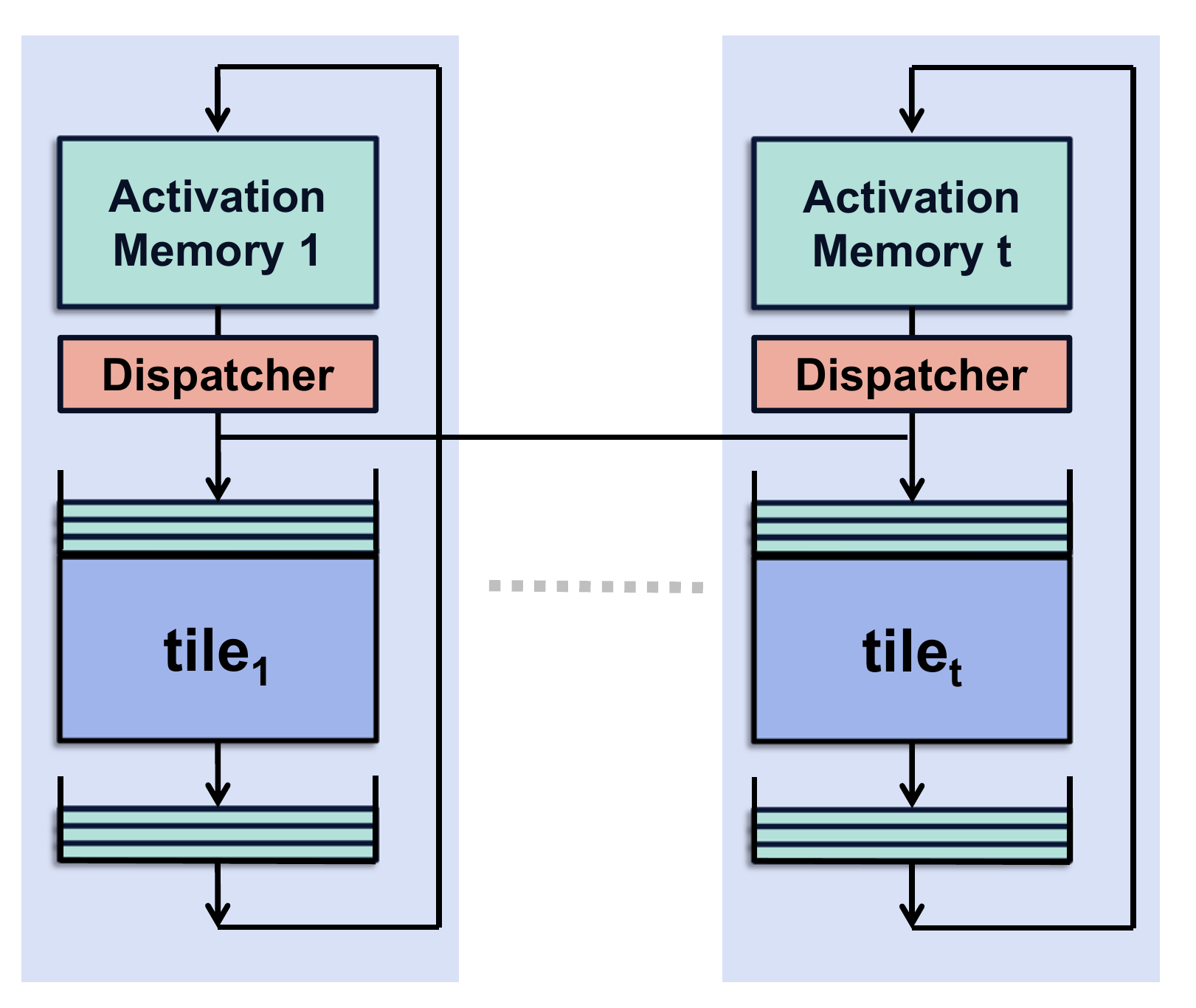}
\caption{A Multi-Tile \OUR. }
\label{fig:i:full}
\end{figure}

Figure~\ref{fig:i:full} shows an \OUR architecture with multiple tiles. Each tile has its own local slice of the AM, a local WM (not shown), an input activation buffer and an output activation buffer. A dispatcher per AM slice reads the appropriate activations as in \STR or \PRA while also accommodating the lookahead needs of \OURS. To reduce energy and bandwidth requirements, the dispatcher uses profile-derived per layer precisions to read only as many activation bits as necessary from AM~\cite{judd:reduced}. Prior to broadcasting each activation group to all tiles, it trims the activations further by dynamically detecting the precision necessary given their values~\cite{dynamicstripes}.
 Locally then each tile consumes their activations at their own pace. The buffers (same as in Figure~\ref{fig:wsi:asu}) determine how far apart the tiles can be in the activation space.  Locally, each tile can perform further processing, for example for \OURPRA we can do encoding to powers of two.

\section{Evaluation}
\label{sec:e}
\begin{table}
\centering\small
\begin{tabular}{|r|r|}
\hline
\textbf{Parameter} & \textbf{Value} \\ \hline
Weight Lanes Per Filter & 16 \\
Filters Per Tile & 16 \\
Tiles & 4 \\
Lookahead & 0-7 \\
Lookaside & 0-6 \\
WM Per Tile & 2MB \\
AM & 4MB (total) \\
Precision & 16-bit \\
Per Tile Activation Buffer & 2KB \\
\hline
\end{tabular}
\caption{Baseline and Base \OURS configurations.}
\end{table}
A cycle level simulator is used to evaluate the relative performance of \OURS and its variants by modeling execution time for convolution and fully connected layers. Table~\ref{nets} lists the pruned CNNs studied here and their \textit{static} and \textit{effective} weight sparsity which are respectively the ratios $\frac{Zero\ Weights}{Total\ Weights}$ and $\frac{Multiplies\ with\ Zero\ Weight}{Total\ Multiplies}$. The effective weight sparsity is a more relevant metric as it is directly proportional to performance potential.
All area and energy measurements were performed over layout using circuit activity for representative data inputs. The layouts were generated for a TMSC 65nm technology using Cadence Innovus after synthesizing them with Synopsys Design Compiler. We used the typical case design library as it yields more pessimistic results for our designs. All designs operate at 980MHz and the results are normalized against the DCNN accelerator design detailed in Table~\ref{tbl:base:config}. SRAMs were modeled via CACTI~\cite{Muralimanohar_cacti6.0} and eDRAM via Destiny~\cite{destiny}.
\begin{table*}
\centering
\small
\begin{tabular}{|r|r|r|r|}
\hline
\textbf{} & \textbf{Static Weight} & \textbf{Effective Weight}& \textbf{} \\
\textbf{Network} & \textbf{Sparsity (\%)} & \textbf{Sparsity (\%)}&\textbf{Acronym} \\ \hline
AlexNet-Eyeriss~\cite{cvpr_2017_yang_energy} & 90.6 & 86.4 & AlexNet-ES\\ \hline

AlexNet-SkimCaffe~\cite{SkimCaffePaper} & 87.5 & 79.4 & AlexNetS-SS\\ \hline
GoogLeNet-Eyeriss~\cite{cvpr_2017_yang_energy} & 66.1 & 74.8 & GoogLeNet-ES\\ \hline
GoogLeNet-SkimCaffe~\cite{SkimCaffePaper} & 78.0 & 61.6 & GoogLeNet-SS\\ \hline
ResNet-50-SkimCaffe~\cite{SkimCaffePaper} & 50.4 & 44.9 & ResNet50-SS\\ \hline

\end{tabular}
\caption{Networks Studied.}
\label{nets}
\end{table*}

\newcolumntype{P}{>{\centering\arraybackslash}m{1.8cm}}
\begin{table*}
\small
\centering
\begin{tabular}{|r|P|P|P|r|r|r|}
\hline
\textbf{Accelerator} & \textbf{Configuration} & \textbf{Performance} & \textbf{Power} & \textbf{Area} & \textbf{Frequency} & \textbf{Tech. Node}\\ \hline
\textbf{\BASE~\cite{DaDiannao}} & 4-16-16 & 1\ Tmul/sec& $6.94\ Watt$& $29.68mm^2$ & 980 Mhz & 65nm\\ \hline
\end{tabular}\\
\caption{Baseline Accelerator. Configuration is labeled as ``Tiles - Filters/Tile - Products/Filter''.}
\label{tbl:base:config}
\end{table*}

The rest of this section is organized as follows: Section~\ref{sec:e:wsp} reports performance improvements when \OURS skips weights only and identifies a few candidate configurations for further exploration. Sections~\ref{sec:e:a} evaluates the performance, area, and energy efficiency of \OURPRA and \OURSTR. Section~\ref{sec:e:scnn} compares with a state-of-the-art sparse accelerator. Finally, Section~\ref{sec:e:la:other} shows that it is possible to further improve performance by not restricting the lookaside connections to one step ahead.

\subsection{Weight Sparsity}
\label{sec:e:wsp}
Here we evaluate \OURS when it exploits \textit{only} weight sparsity. We explore a spectrum of design points with varying lookahead and lookaside, and compare their performance to the \BASE architecture.
We consider combinations of lookahead $h$ and lookaside $d$ such that $h + d - 1 = 2^{n}$, and $n = \{2, 4, 8\}$. We restrict attention to designs where the lookaside connections are to exactly one step ahead. Section~\ref{sec:e:la:other} will show that further benefits are achievable when this restriction is relaxed.

\begin{figure}
\centering
\includegraphics[width=0.5\textwidth]{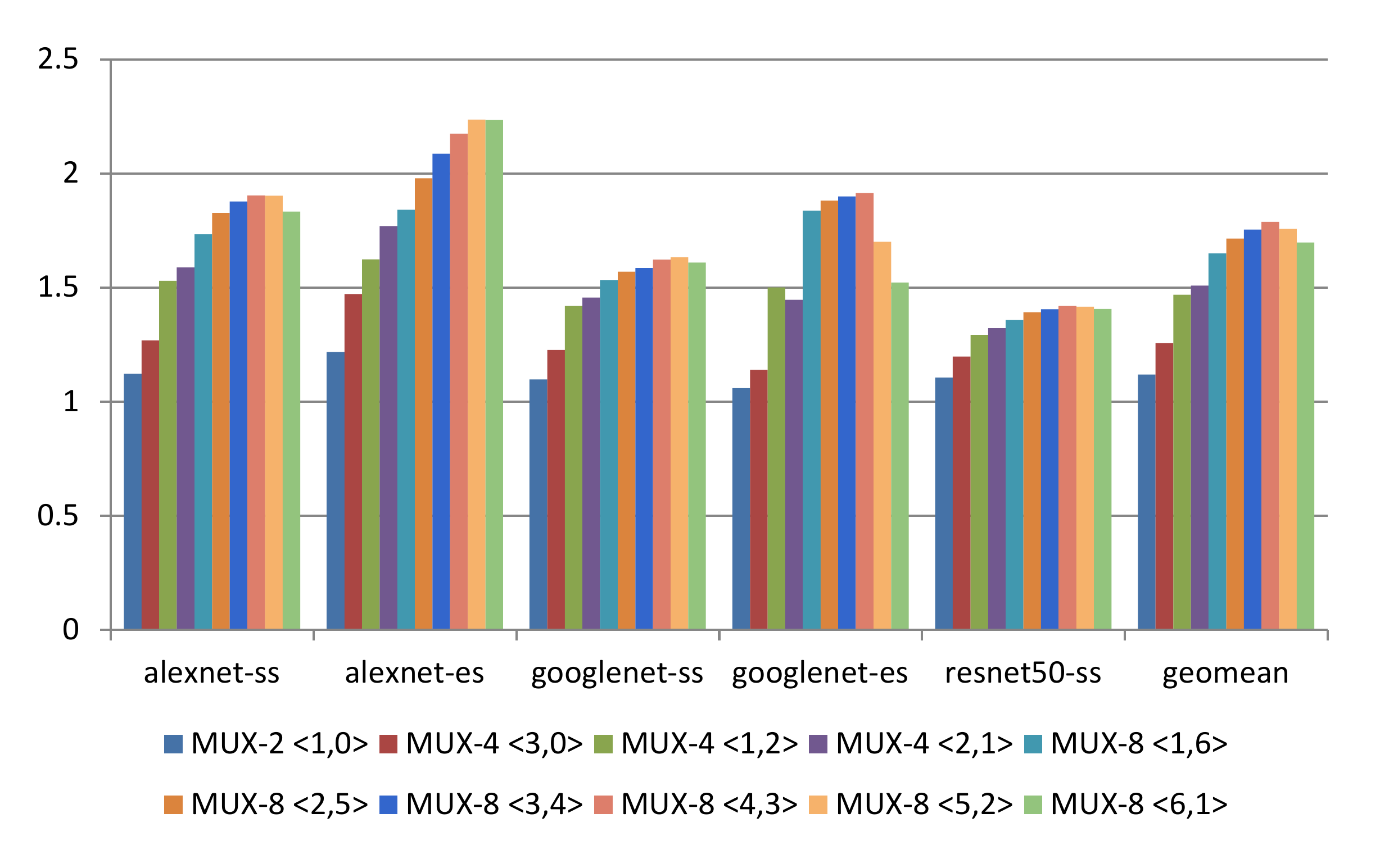}
\caption{Per-network lookahead and lookaside effect on performance.}
\label{fig:exploration}
\end{figure}

Figure~\ref{fig:exploration} reports speedup vs. \BASE where configurations are annotated as ``\texttt{MUX-n} $\langle h,d \rangle$''.
Using a larger multiplexer as expected always results in better performance regardless of the lookahead and lookaside mix. For a fixed multiplexer input size adding a small number of lookaside inputs by ``sacrificing'' lookahead inputs offers a significant marginal gain in performance.
For example, this can be seen in the transition from \texttt{MUX-4} $\langle 3,0 \rangle$ to either \texttt{MUX-4} $\langle 1,2 \rangle$, or \texttt{MUX-4} $\langle 2,1 \rangle$. Lookaside is the
only mechanism that allows a weight lane heavily populated with
effectual weights to distribute its load to neighboring ones,
thus reducing weight lane imbalance. Yet, it is generally unlikely for some weight lane to have multiple adjacent ``heavy'' lanes. Thus, arbitrarily expanding the lookaside window yields diminishing returns. Similarly, adding large lookahead will impact the effectiveness of the activation handling back-end as per Section~\ref{sec:full}.

\sloppy
Performance with AlexNet-ES is generally higher than with AlexNet-SS. The weight pruning method used in AlexNet-ES is distribution-aware which translates to less imbalance across weight lanes in \OURS. This result motivates work on \OURS-aware pruning methods and in improving the statically generated weight schedule.

In general, even when most weights are zero, the lookahead $h$ limits the maximum possible speedup from weight skipping to $(h+1)\times$. Taking into account the weight sparsity of the studied networks, and the potential that exists for any given lookahead choice, the results show that \OURS captures a significant potion of this potential. For example, \texttt{MUX-8}~$\langle 2,5 \rangle$  improves performance with resnet50-SS by nearly $1.4X$, a respectable showing given that this network's effective weight sparsity is about 45\% and that the maximum speedup possible, if all weights were zero, would have been $3\times$.

Since adding lookaside and/or lookahead capability incurs area overheads due to multiplexer size, variability in wiring, and the need for more ABRs and wider activation selection multiplexers in the ASU, any design choice must take this trade-off into account. The results show that there is a wide selection of choices that a designer can make to tune their design for their specific goals. In the interest of space and clarity, based on area estimates of a \OURPRA tile with varying $\langle h,d \rangle$ settings through synthesis and layout the rest of the evaluation restricts attention to the following three configurations:
\texttt{MUX-8} $\langle 1,6 \rangle$,
\texttt{MUX-8} $\langle 2,5 \rangle$, and \texttt{MUX-8} $\langle 4,3 \rangle$. For clarity, we will identify them with ``$\langle h,d \rangle$'' from this point on.

\subsection{Activations: \OURPRAcore\ and \OURSTRcore}
\label{sec:e:a}
\newcommand{\OURCONF}[2]{$\langle #1, #2 \rangle$}
\newcommand{\OURPRAC}[2]{\textit{\OURPRAcore}\OURCONF{#1}{#2}}
\newcommand{\OURSTRC}[2]{\textit{\OURSTRcore}\OURCONF{#1}{#2}}
\subsubsection{Performance}
\label{sec:e:a:p}

\begin{figure}
\centering
\includegraphics[width=0.5\textwidth]{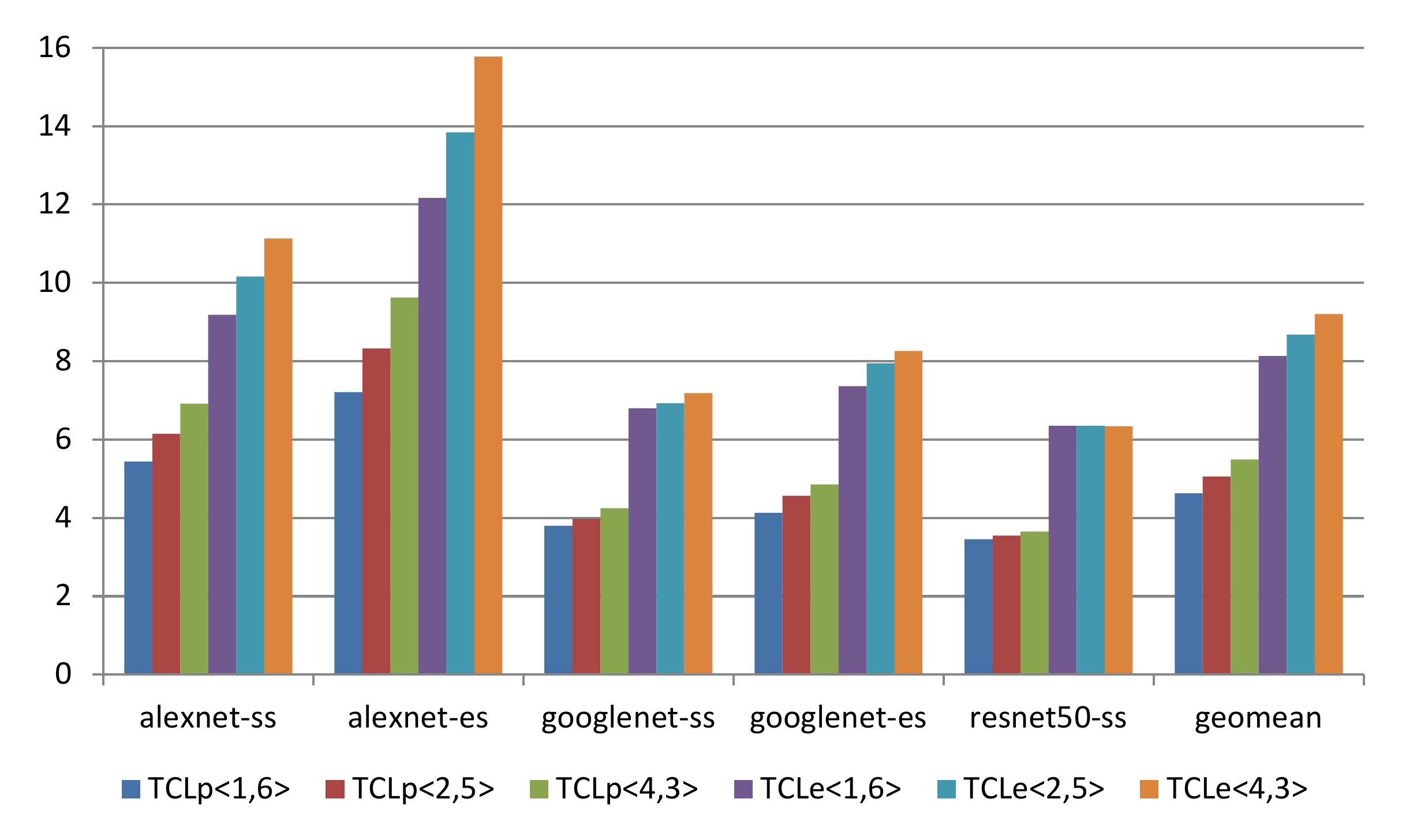}
\caption{Performance with \OURPRAcore\ and \OURSTRcore.}
\label{fig:e:a:perf}
\end{figure}

Figure~\ref{fig:e:a:perf} reports the relative to \BASE performance of several \OURPRA and \OURSTR configurations and for all layers. Adding the capability to exploit precision variability mostly complements ineffectual weight skipping. On average, even the least capable \OURSTRC{1}{6} (as per the results of Section~\ref{sec:e:wsp}) improves performance by \OURSTRlowperf over \BASE. The benefits are much higher for the Alexnet variants than for Googlenet-SS and ResNet50-SS. Alexnet has only a few layers which are larger by comparison to the other two networks. As a result, the units are better utilized.
For these results we do not restrict the rate at which activations can be read from AM. We do so to demonstrate the potential of \OURPRA and \OURSTR. Depending on the application needs, the AM and the interconnect have to be designed to balance cost vs. the desired performance. Given that the access pattern is regular and strided, banking and distributing AM should be sufficient to sustain the improved processing rate.

All \OURPRA configurations outperform any \OURSTR configuration which is expected since the potential from exploiting ineffectual activation bit content~\cite{Pragmatic} is much higher than that of exploiting precision~\cite{Stripes-MICRO,dynamicstripes}. However, there overall performance benefit is lower than what the ineffectual bit content would suggest. Cross activation lane  synchronization is the culprit here as all activation lanes within a lookahead window have to wait for the activation lane with the most oneffsets to finish before advancing to the next lookahead window. This is more pronounced for ResNet50-SS where increasing lookahead results in less performance benefits.

This also explains why \OURPRAC{1}{6} the \OURPRA configuration with the least lookahead generally outperforms \OURSTRC{4}{3}.
overall, while \OURPRA generally outperforms \OURSTR, it is a more expensive design as it requires more wires per activation than \OURSTR. The next two sections report their relative area and energy efficiency. A designer can take into account this trade off to decide which design is best suited for their application.

\subsubsection{Energy Efficiency}
\label{sec:e:a:e}
\begin{figure}
\centering
\includegraphics[width=0.5\textwidth]{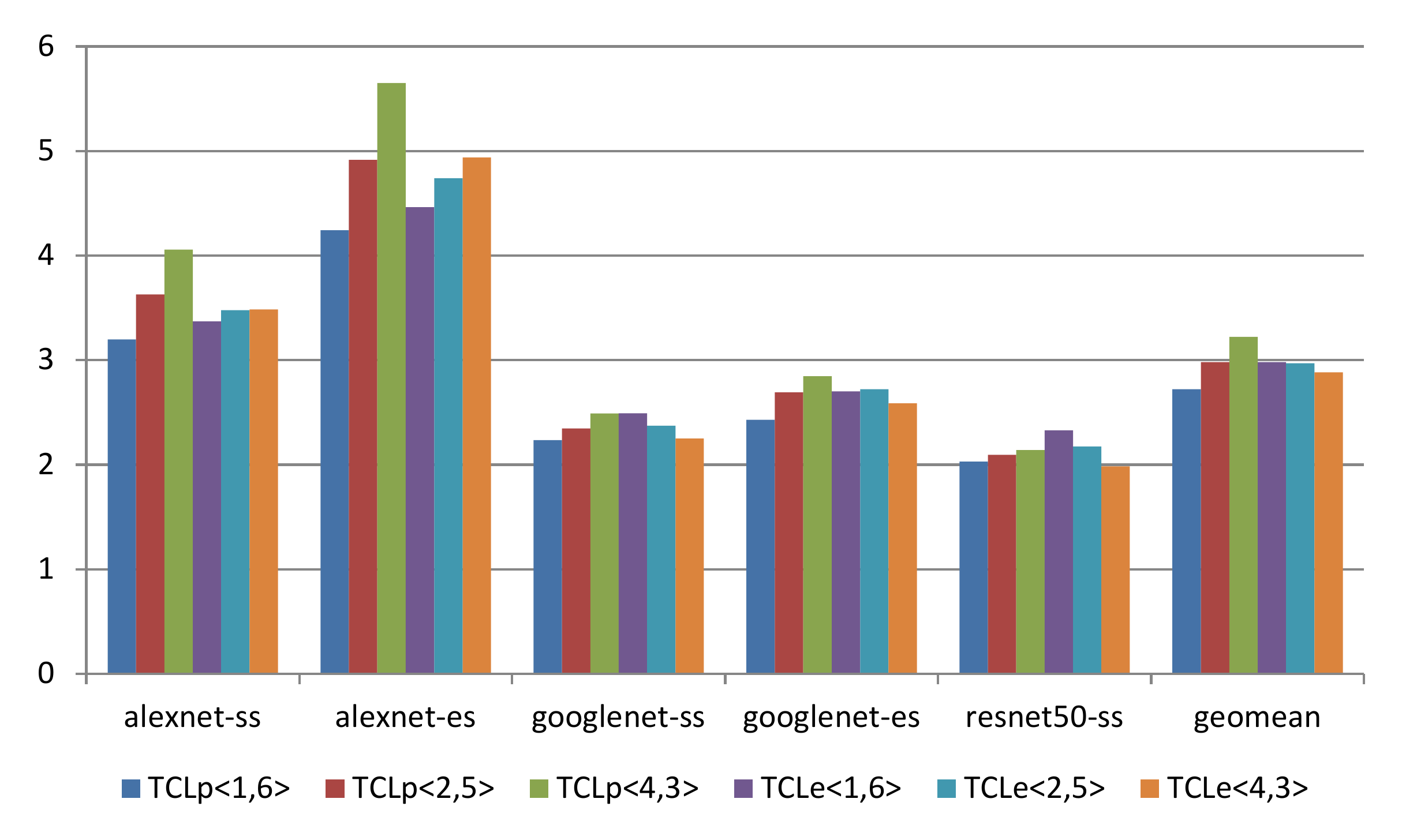}
\caption{Energy Efficiency with \OURPRAcore\ and \OURSTRcore.}
\label{fig:e:a:ee}
\end{figure}

Figure~\ref{fig:e:a:ee} reports the energy efficiency of the \OURPRA and \OURSTR configurations relative to \BASE. All designs prove more energy efficient as the performance benefit far outweighs the additional hardware power cost. \OURSTR is generally more energy efficient than \OURPRA which is expected as it is a lower cost design. Across the three configurations \OURSTR is 2.90x more efficient than \BASE, while \OURPRA is only 2.32x more efficient. The most efficient configuration is \OURSTRC{4}{3} with 3.25x relative energy efficiency.

\subsubsection{Area}
\label{sec:e:a:a}
Table~\ref{tbl:area} reports the area for various accelerators. For clarity, we report detailed breakdowns only for \OURPRAC{1}{6}, \OURSTRC{1}{6}, and \BASE.  The area vs. performance trade off is sublinear which suggests that even if performance could scale linearly for \BASE it would still trail in performance per area. In practice performance in \BASE scales sublinearly with area as the typical filter count, and the typical dimensions of the filters, the input and the output  result in higher underutilization for wider configurations of \BASE.

The area differences among the configurations studied are negligible. Since the sum of lookahead and lookaside is the same for all configurations. Overall, most of the area is in the memories and to a lesser extend in the SIPs.

\begin{table*}
\centering
\small
\begin{tabular}{|c|c|c|c|}
\hline
                   & \textbf{\OURPRA<1,6> area ($mm^2$) }& \textbf{\OURSTR<1,6> area ($mm^2$)} & \textbf{DCNN area ($mm^2$)} \\ \hline
\textbf{Compute Core }      & 16.33           & 9.25            & 3.2             \\
\hline
\textbf{Weight Memory}      & 12.03           & 12.03           & 12.03           \\
\hline
\textbf{Act. Input Buffer } & 3.66            & 3.66            & 3.66            \\
\hline
\textbf{Act. Output Buffer} & 3.66            & 3.66            & 3.66            \\
\hline
\textbf{Activation Memory}  & 7.13            & 7.13            & 7.13            \\
\hline
\textbf{Dispatcher}         & 0.37            & 0.39            & -               \\
\hline
\textbf{Offset Generator}   & 0.11            & -               & -               \\
\hline
\textbf{Total}              & 43.29           & 36.12           & 29.68           \\
\hline
\textbf{Normalized Total }  & 1.46            & 1.21            & 1.00           \\
\hline\hline
                  & \textbf{\OURPRA<2,5> area ($mm^2$)} & \textbf{\OURSTR<2,5> area ($mm^2$)} & \textbf{DCNN area ($mm^2$)} \\ \hline
\textbf{Normalized Total }  & 1.47            & 1.22            & 1.00  \\ \hline
\hline
                  & \textbf{\OURPRA<4,3> area ($mm^2$)} & \textbf{\OURSTR<4,3> area ($mm^2$)} & \textbf{DCNN area ($mm^2$)} \\ \hline
                  \textbf{Normalized Total}   & 1.48            & 1.23            & 1.00 \\\hline
\end{tabular}
\caption{Area Breakdown for \OURPRA and \OURSTR.}
\label{tbl:area}
\end{table*}

\subsection{Comparison With a Sparse Accelerator}
\label{sec:e:scnn}

To compare our design with an existing sparse accelerator on the same networks we implement a cycle level performance model of SCNN~\cite{SCNN} with the same number of multipliers (1024) as \OURS.
This design partitions activation feature maps in the X and Y dimensions and distributes them to an $8\times 8$ array of processing elements (PEs). Each PE holds a copy of the weights and computes the local convolution with 16 multipliers to perform a $4\times 4$ Cartesian product per cycle. Non-zero activations and weights are fed in with their corresponding coordinates and the resulting products are mapped to 32 accumulator banks using a 16 to 32 crossbar. In the case where two products map to the same accumulator the pipeline incurs a stall.
The activation partitioning requires that PEs exchange partial sums to compute the overlapping activations. We assume each PE can send one partial sum to each of its neighbors per cycle. For layers using a non-unit stride we partition the weights and activations according to the mutually exclusive groups of outputs that they produce. We assume sufficient off-chip bandwidth to keep the units busy. Given that SCNN's peak compute bandwidth for fully-connected layers is only 1/4 of that for convolutional layers, here we limit attention only to convolutional layers.

\begin{figure}
\centering
\includegraphics[width=0.5\textwidth]{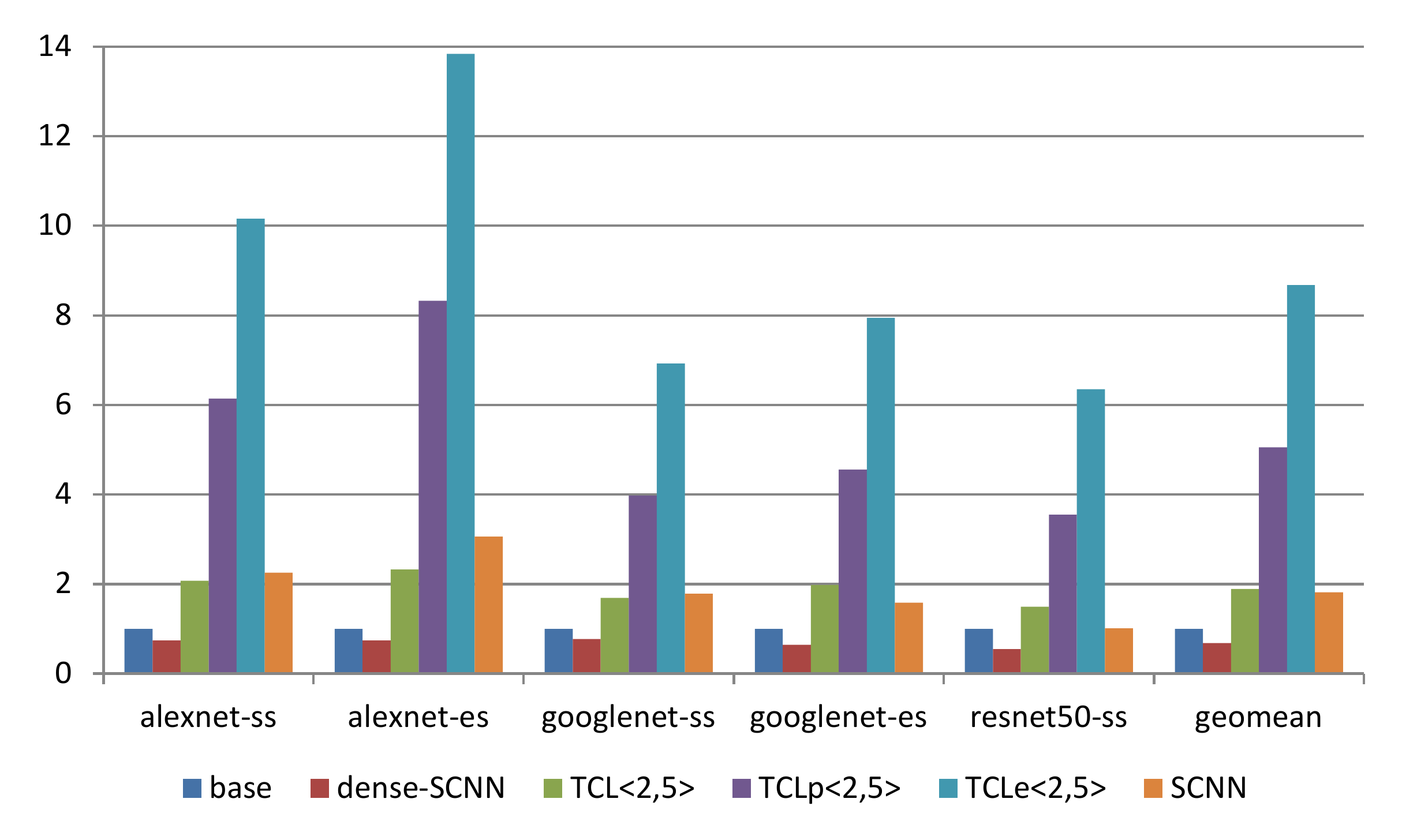}
\caption{Comparing with a state-of-the-art sparse accelerator: Relative Performance.}
\label{fig:e:scnn}
\end{figure}

Figure~\ref{fig:e:scnn} reports the relative performance of SCNN, weight-skipping only \OURS, \OURSTR and \OURPRA for the \OURCONF{2}{5} configuration. Dense-SCNN shows SCNN running on the dense version of each network. On average SCNN on dense networks is \SCNNvsOURSdense than our baseline inner product based accelerator. This agrees with their observation that inner product accelerators are better suited to dense networks.
One exception is Resnet50, where SCNN in \SCNNvsOURSresnet. This is due to their mapping of the activations
to the PE array where activations are partitioned in the x and y dimensions. For example the later layers have dimensions $(A_x,A_Y) = (7,7)$ so only one $(x,y)$ is mapped to each PE and the Cartesian products must be computed one channel at a time.
When we exploit precision and effectual bit content in addition to weight sparsity in \OURSTR and \OURPRA, we achieve a speedup over SCNN of \SCNNvsOURSTR and \SCNNvsOURPRA, respectively.

\begin{table*}
\centering
\small
\begin{tabular}{|l|c|c|c|c|l|l|}
\hline
\textbf{}            & \textbf{\begin{tabular}[c]{@{}c@{}}Skip\\ MACC\end{tabular}} & \textbf{\begin{tabular}[c]{@{}c@{}}Skip\\ Memory \\ Access\end{tabular}} & \textbf{\begin{tabular}[c]{@{}c@{}}Reduced\\ MACC\end{tabular}} & \textbf{\begin{tabular}[c]{@{}c@{}}Reduced\\ Memory\\ Access\end{tabular}} & \multicolumn{1}{c|}{\textbf{\begin{tabular}[c]{@{}c@{}}Data Routing\\ Type \&\\ Mechanism\end{tabular}}}                             & \multicolumn{1}{c|}{\textbf{Inner Spatial Dataflow}}            \\ \hline
\textbf{Cnvlutin}    & IA                                                           & IA                                                                       & -                                                               & -                                                                          & \begin{tabular}[c]{@{}l@{}}Weight-Dynamic/Activation-Static\\ Sparse at Input: Independent Weight Ports\end{tabular}                 & \begin{tabular}[c]{@{}l@{}}Dot Product\\ Reduction\end{tabular} \\ \hline
\textbf{Cambricon-X} & IW                                                           & IW                                                                       & -                                                               & -                                                                          & \begin{tabular}[c]{@{}l@{}}Weight-Static/Activation-Dynamic\\ Dense at Input: Activation Crossbar\end{tabular}                       & \begin{tabular}[c]{@{}l@{}}Dot Product\\ Reduction\end{tabular} \\ \hline
\textbf{SCNN}        & IA+IW                                                        & IA+IW                                                                    & -                                                               & -                                                                          & \begin{tabular}[c]{@{}l@{}}Weight-Dynamic/Activation-Dynamic\\ Dense at Output: Product Crossbar\end{tabular}                        & Cartesian Product                                               \\ \hline
\textbf{\OURPRA/\OURSTR}   & IW                                                           & IW                                                                       & IA+EA                                                           & IA+EA                                                                      & \begin{tabular}[c]{@{}l@{}}Weight-Static/Activation-Dynamic\\ Sparse At Input: Sparse Shuffling Network for Activations\end{tabular} & \begin{tabular}[c]{@{}l@{}}Dot Product\\ Reduction\end{tabular} \\ \hline
\end{tabular}
\caption{Qualitative Comparison of CNN Accelerators.}
\label{tbl:qcmp}
\end{table*}

\subsection{Alternate Lookaside Promotion Patterns}
\label{sec:e:la:other}
Thus far, we have assumed lookahead and lookaside patterns must constitute a contiguous window in the time and lane directions, respectively. However, weight promotions need not be restricted to continuous vectors in these orthogonal directions only, but may come from an arbitrary coordinate that is a combination of both lookahead and lookaside. That is, given a lookahead distance of $h$, it is possible to implement a lookaside pattern that allows promotion from any subset of the $16 \times h$ positions in this window (where 16 is the filter lane width). Here, we evaluate a \textit{sparse promotion pattern} that allows weight promotion from arbitrary locations in the weight stream. The term ``sparse'' here refers to the fact that a weight $w[lane, step]$ that can steal from location $[lane+d, step+h]$ may not necessarily have a connection to steal from locations $[lane+d-1, step+h]$ or $[lane+d, step+h-1]$.

To simplify the search space, we consider a lookahead $h=2$, a lookaside $d=5$, and limit lookaside to a distance of 7 lanes, for a total of $2 + 2 \times 7 = $ 16 possible promotion connections. Due to power and area considerations, we limit our study to connectivity that results in $h+d=7$ \textit{promotion sites} resulting in an 8-input multiplexer (the extra input is for no promotion).

To perform a search in this space, we begin with all 16 promotion sites connected and iteratively remove the connection that, when removed, results in the smallest performance degradation for a given network. This results in a connectivity pattern tailored to each network that gives improved performance over the $\langle 2,5 \rangle$ design that uses the same amount of lookahead (therefore having similar area requirements). Given that the connectivity pattern is fixed in hardware, we also report results for a fixed checkerboard-like pattern (``\textit{Checkers}'') that achieves most of the speedup of the per-network optimized patterns. Results for LA-2 Max, a design that allows promotion from any location within a filter lane and within a lookahead of 2, are shown for comparison.

\begin{figure}
\centering
\includegraphics[width=0.5\textwidth]{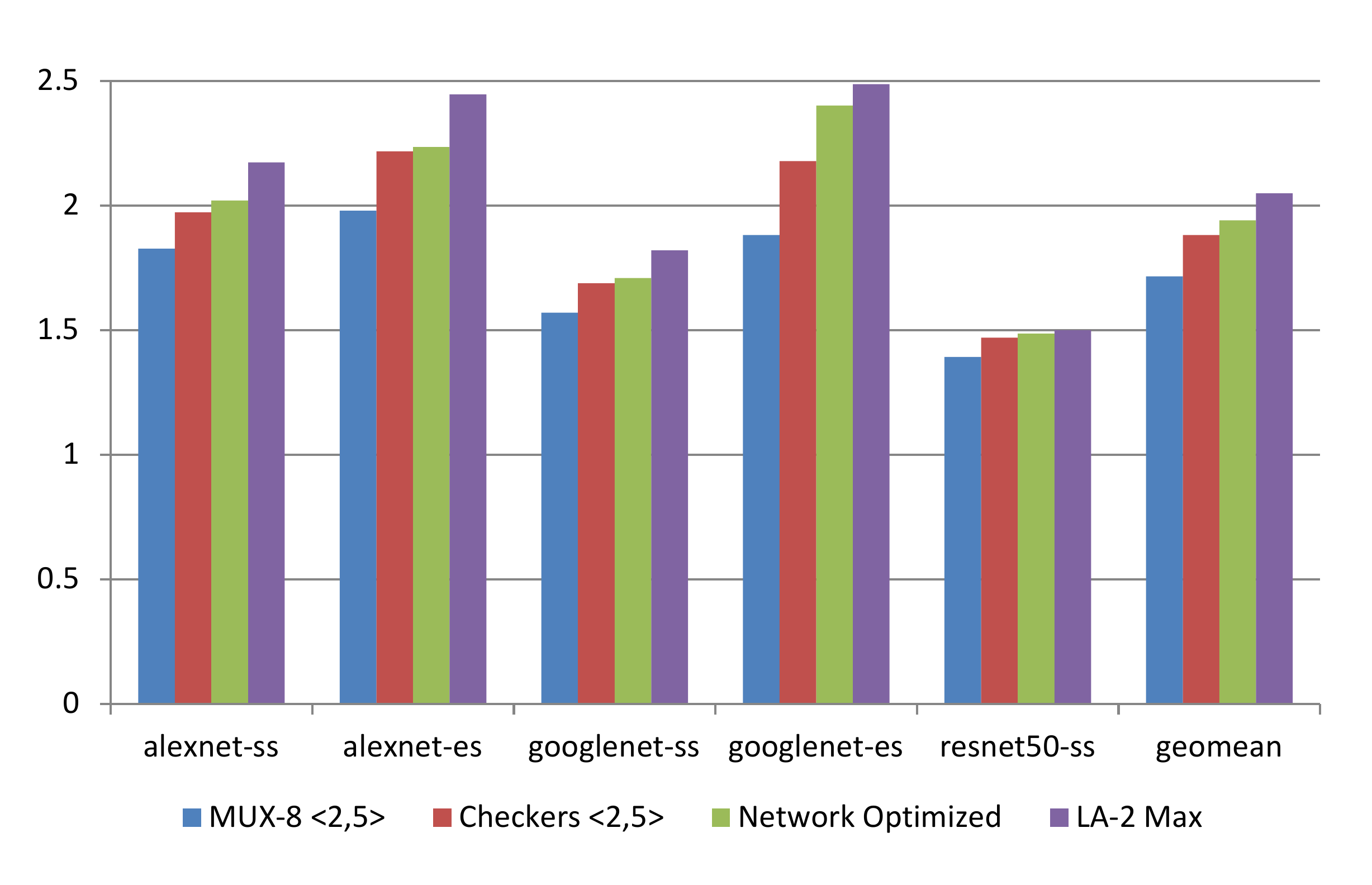}
\caption{Performance with a sparse promotion window for \OURScore.}
\label{fig:e:a:sparse}
\end{figure}

The network-optimized sparse promotion patterns can offer up to a 27\% performance increase (with GoogLeNet-Eyeriss) over the standard $\langle 2,5 \rangle$ design, without necessitating any additional lookahead and associated overhead.

\section{Related Work}
\label{sec:related}
Recent CNN hardware accelerators exploit properties of the weights and of the activations to boost performance and energy efficiency beyond what is possible by mapping the CNN computation flow alone onto hardware. These properties include ineffectual weights and activations~\cite{han_eie:isca_2016, albericio:cnvlutin, cambricon:2016,SCNN}, precision reduction,e.g.,~\cite{binaryconnect,quantizedBlog}, precision variability~\cite{Stripes-MICRO}, and effectual bit content~\cite{Pragmatic}. Due to space limitations we restrict attention to accelerators that exploit weight and activation sparsity. Section~\ref{sec:ia:whynot} has discussed how skipping ineffectual activations relates to precision reduction and ineffectual activation bit skipping. Here we review specific designs. Table~\ref{tbl:qcmp} highlights the following most relevant characteristics of these designs: 1)~for which input data it skips the multiply-accumulate computation, 2)~for which input data it avoids a memory reference, 3)~for which input data it performs a reduced cost multiply-accumulate, 4)~for which input data it performs a reduced cost memory access, 5)~how is the input data routed to the appropriate compute unit or storage unit, and 6)~the ordering used to compute inner-products.

Cnvlutin skips both the computation and the memory access for ineffectual activations (IA). It requires no special input or output routing mechanism other than independent weight ports per group of weights that pair up with each activation~\cite{albericio:cnvlutin}.

Cambricon-X exploits ineffectual weights (IW) in an inner product based accelerator~\cite{CambriconX}. Non-zero weights are compacted in memory and tagged with deltas (distance between weights). Each cycle one PE (equivalent to our inner product unit) fetches 16 weights and selects the corresponding 16 activations from a vector of 256. Chained adders are used to decode the deltas into absolute offsets. It uses a 256-wide \textit{input activation crossbar} to pair up activations with the corresponding weights. This approach is similar to \OURS with a very large 16x16 lookahead window and encoded mux selects. This requires a memory interface for 256 activations, 16 times that of DianNao~\cite{diannao}.
The authors discuss that this activation bandwidth makes their approach impractical for scalable accelerators like DaDianNao~\cite{DaDiannao}.

SCNN skips computations and memory accesses for both ineffectual weights and ineffectual activations~\cite{SCNN}. It compacts weights and activations in memory where only the effectual elements are stored each followed by the number of ineffectual elements that have been omitted. A $16\times32$ \textit{output crossbar} routes multiplication results to 32 accumulator banks. SCNN is designed to minimize input read bandwidth. Since SCNN uses 4x4 Cartesian Products it is only able to use 4 of the 16 multipliers for FCLs which have no weight reuse.

\OURS skips computations and memory accesses for ineffectual weights albeit to a different degree than SCNN or Cambricon-X. It reduces the bandwidth and energy cost of the memory accesses for both ineffectual and effectual activations (EA). It matches activations and weights using a hybrid \textit{input weight-static/activation-dynamic} approach since it utilizes a \textit{sparse shuffling network} for the input activations and restricted static scheduling for the weights.

To capture sparsity, SCNN and Cambricon-X use a \textit{dense} hardware interconnect. SCNN uses an output crossbar whereas Cambricon-X uses an input crossbar. \OURS uses a \textit{sparse} input interconnect to capture a sufficient number of ineffectual weights and compensates for the loss in opportunity by targeting all activations instead.

\section{Limitations}
\label{sec:limitations}
We presented an implementation of \OURS that builds upon the vector-like approach of DaDianNao. Other accelerators, such as Eyeriss, SCNN, or the TPU favor instead a grid-like PE organization. It would be interesting to further consider a grid-like design that uses the same principles as \OURS to exploit its different combination of ineffectual computations. However, we believe that this work stands on its own for the following two reasons: First, it demonstrates that a particular combination of ineffectual computations offers greater potential than past considered options. Second, it proves that it is possible to exploit this combination in practice via a practical design. For these reasons, this work motivates  studying how \OURS's approach interacts with grid-like designs. We believe that \OURS approach is compatible with such designs.

We assumed that all the data for each layer can fit on-chip using appropriately sized eDRAMs. Other designs such as the TPU or DaDianNao employ large on-chip eDRAMs. However , depending on the network and the design constraints it will not always be possible to fit all data on chip. For example, for image classification networks, the inputs are likely to grow larger following the trend toward higher resolution image sensors. Accordingly, these observations point to two follow up studies. The first would be to revise the on-chip memory hierarchy to utilize either additional levels, such as SRAM buffers for the weights and activations. Past work has shown that it is possible to effectively use the significant reuse in the data stream to greatly reduce the need for off-chip accesses~\cite{EyerissISCA2016,Gao:Tetris:ASPLOS}. The second, potentially complementary study, would need to look at how to block the processing of larger networks when the on-chip storage space is limited, e.g.,~\cite{TPUISCA17,Gao:Tetris:ASPLOS}. Nevertheless, we believe that this work again serves as the motivation for these follow up studies.

While we have shown that \OURS can outperform an SCNN that has the same peak computation bandwidth, we did not study in detail how the overall energy consumption compares between these two implementations. Presently this study is not directly possible because it would require revisiting the memory hierarchy design for \OURS as described earlier in this section. This is necessary as SCNN uses relatively small buffers to hold its data. It would also require revisiting the way activations and weights are partitioned and scheduled across tiles. However, we believe that \OURS is not fundamentally incompatible with the grid-like approach of SCNN nor with the use of a multilevel memory hierarchy. Moreover, while SCNN is optimized for reducing data movement especially for the inputs, it has to route each individual product to an accumulator bank either within the PE or to neighboring PEs. Moreover, filters are replicated across tiles as necessary.  In\OURS activations are presently broadcast to all tiles since the filters are not replicated. Moreover, several products are reduced together in the adder trees. Hence there are fewer results of a multiply-accumulate that are being communicated. At the weight inputs, \OURS still processes some weights that are zero. However, the overall input bandwidth for weights should be reduced proportionally to the speedup gained by weight skipping alone. Accordingly, the relative overhead would be lower compared to a dense network. Furthermore, by using precision it should be possible to reduce the storage requirements and bandwidth for both weights and activation. As with the other limitations discussed earlier, we believe that this study serves as motivation for future work to addresse them.

\section{Conclusion}
\label{sec:theend}
We showed \textit{what} combination of input value properties is better to exploit and \textit{how} to do so with a practical design. We believe that \OURS is a compelling design for hardware acceleration of CNNs for two reasons:
\begin{itemize}
\item \OURS \textit{does not require any changes to the CNN} to deliver benefits. It offers immediate benefits for CNNs, sparse or not, as long as they exhibit dynamic precision variability or ineffectual activation bits. CNNs are expected to naturally exhibit both these properties since they are a byproduct of how CNNs work at the fundamental level: activations typically follow a distribution where most values cluster near zero with only a few spiking for a given input. While \OURS does not attack ineffectual activations head on, Section~\ref{sec:ia:whynot} explains that by attacking dynamic precision variability and ineffectual bit content, \OURS delivers benefits for both ineffectual and effectual activations alike. A key insight is that in \OURS the opportunity cost of not completely eliminating ineffectual activations is only a fraction of that incurred by bit-parallel accelerators. Targeting other activation values properties for all activations results in benefits that exceed the opportunity lost from not completely eliminating the ineffectual activations.

\item \OURS \textit{rewards CNN designers for changing the CNN}. Specifically, it rewards  optimizations that increase weight sparsity, reduce activation precision, or increase zero bit content for activations. \OURS will ``incentivize'' CNN designers and researchers to further refine such techniques by offering gradual yet immediate benefits for any such advances. For this reason, \OURS can accelerate innovation along the aforementioned design fronts. Eventually, if these efforts mature to a sufficient degree, more specialized hardware designs can take over.
\end{itemize}

While in this work we limited attention to designs with large on-chip memories, future work should consider designs with limited on-chip storage. We also ignored opportunities for further reducing storage for weights such as exploiting statically derived precisions for weights~\cite{Judd:Proteus:ICS}. Additionally, we did not attempt to optimize the weight schedule statically. These are all interesting directions for future work. We believe that \OURS represents a valuable alternative to existing options and that it will motivate further innovation in hardware accelerators designs for CNNs and DNNs.

\bibliographystyle{ieeetr}
\bibliography{ref2}

\end{document}